\def\year{2019}\relax
\documentclass[letterpaper]{article} 
\usepackage{aaai19}  
\usepackage{times}  
\usepackage{helvet}  
\usepackage{courier}  
\usepackage{url}  
\usepackage{graphicx}  

\usepackage{pdfpages}

\def\eqlabel#1{\label{eqn:#1}}

\usepackage{adjustbox}
\usepackage{amsmath,amssymb,amsthm}
\usepackage{mathabx}
\usepackage{multirow}
\usepackage{url}
\usepackage{hhline}
\usepackage{tikz,pgfplots}
\usepackage{xcolor}
\usetikzlibrary{arrows,intersections}
\usepackage{multirow}
\usepackage{pgfplots}
\usepackage{algorithm}
\usepackage[noend]{algpseudocode}

\usepackage{subcaption}
\usepackage{multirow}
\include{plots}
\definecolor{bblue}{HTML}{4F81BD}
\definecolor{rred}{HTML}{C0504D}
\definecolor{ggreen}{HTML}{9BBB59}
\definecolor{ppurple}{HTML}{9F4C7C}
\definecolor{Dark scarlet}{HTML}{560319}
\definecolor{Forest green}{HTML}{1E4D2B}

\newcounter{notecounter}
\newcommand{\enotesoff}{\long\gdef\enote##1##2{}}

\enotesoff

\begin{document}
%
\title{Document Informed Neural Autoregressive Topic Models with Distributional Prior}

\newcommand*{\affaddr}[1]{#1} 
\newcommand*{\affmark}[1][*]{\textsuperscript{#1}}
\newcommand*{\email}[1]{\texttt{#1}}

\author{Pankaj Gupta\affmark[1,2], Yatin Chaudhary\affmark[1], Florian Buettner\affmark[1], Hinrich Sch\"{u}tze\affmark[2]\\ 
 \affaddr{\affmark[1]Corporate Technology, Machine-Intelligence (MIC-DE), Siemens AG  Munich, Germany}\\
  \affaddr{\affmark[2]CIS, University of Munich (LMU) Munich, Germany} \\
  {\tt \{pankaj.gupta, yatin.chaudhary, buettner.florian\}@siemens.com}\\
  {\tt pankaj.gupta@campus.lmu.de |  inquiries@cislmu.org}
}

\maketitle

\begin{abstract}

We address two challenges in topic models: 
(1) Context information around words helps in determining
their actual meaning, e.g., ``networks'' used 
in the contexts {\it artificial neural networks} vs.\ {\it biological neuron networks}. 
Generative topic models infer topic-word distributions, taking no or only little context into account. 
Here, we extend a neural autoregressive topic model to exploit the full context information around words 
in a document in a language modeling fashion. The proposed model is named as {\it iDocNADE}. 
(2) Due to the small number of word occurrences (i.e., lack of context) in short text  and data sparsity in a corpus of few documents, the application of topic models is challenging on such texts.  
Therefore,  
we propose
a simple and efficient way of incorporating 
external knowledge 
into neural autoregressive topic models: we use embeddings as
a distributional prior.
The proposed variants are named as {\it DocNADEe} and {\it iDocNADEe}. 

We present novel neural autoregressive topic model variants that consistently outperform state-of-the-art generative topic models 
in terms of generalization, interpretability (topic
coherence) and applicability (retrieval and classification)
over 7 long-text and 8 short-text datasets 
from diverse domains. 

\end{abstract}

\section{Introduction}

Probabilistic topic models, such as LDA \cite{Blei:81},  Replicated Softmax (RSM) \cite{Salakhutdinov:82} and  Document Autoregressive 
Neural Distribution Estimator (DocNADE) 
 \cite{Hugo:82} are often used to extract 
topics from text collections and learn 
document representations to perform NLP tasks such 
as information retrieval (IR), document classification or summarization.

To motivate our first task of {\it incorporating full contextual information}, 
assume that we conduct topic analysis on a collection of research papers from 
NIPS conference, where
one of the popular terms is ``networks". 
However, without 
context information (nearby and/or distant words), its actual meaning is ambiguous since it can refer to such different concepts as 
{\it artificial neural networks} in {\it computer science}  or {\it biological neural networks} in {\it neuroscience} or  {\it Computer}/{\it data networks} in {\it telecommunications}. 
Given the 
context, one can determine the actual meaning of  ``networks", for instance,  
``Extracting rules from artificial neural \underline{networks} with distributed representations", or 
``Spikes from the presynaptic neurons and postsynaptic neurons in small \underline{networks}" or 
``Studies of neurons or \underline{networks} under noise 
in artificial neural \underline{networks}" 
or ``Packet Routing in Dynamically Changing  \underline{Networks}". 
%

\begin{figure*}[t]
  \centering
  \includegraphics[scale=.62]{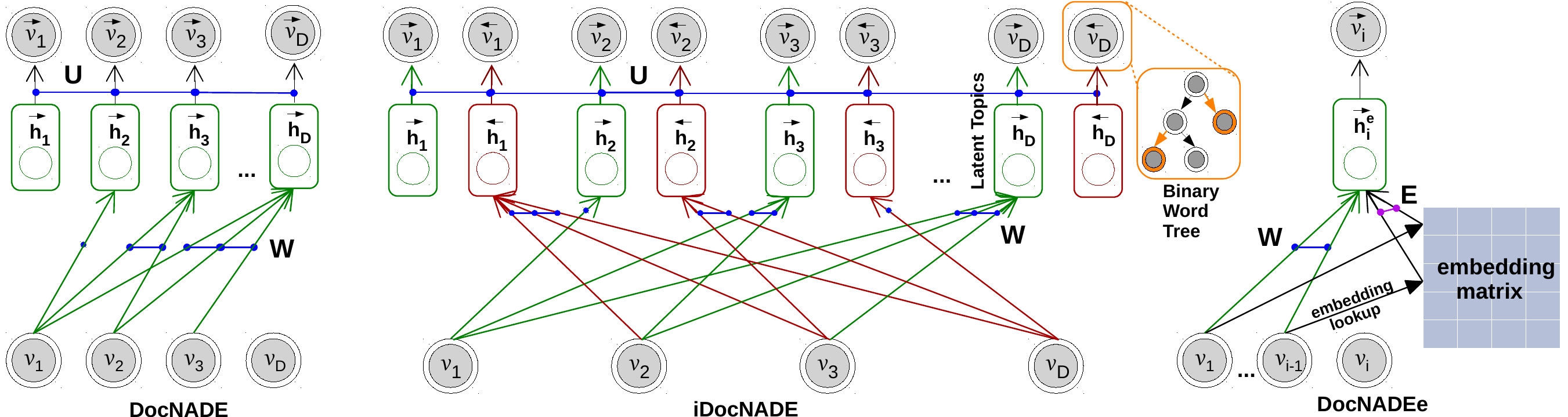}
  \caption{{\it DocNADE} (left), {\it iDocNADE}
  (middle) and  DocNADEe (right) models.  
Blue colored lines signify the connections that share parameters.  
The observations ({\it double circle}) for each word $v_i$ are multinomial. 
Hidden vectors in {\it green} and {\it red} colors identify the forward and backward network layers, respectively.   
Symbols $\overrightarrow{v}_{i}$ and $\overleftarrow{v}_{i}$ 
represent the autoregressive conditionals $p(v_i | {\bf v}_{< i})$ and $p(v_i | {\bf v}_{> i})$, respectively. 
Connections between each $v_i$ and hidden units are  shared, 
and each conditional {$\overrightarrow{v}_{i}$}  (or {$\overleftarrow{v}_{i}$}) is decomposed into a tree of binary logistic regressions, i.e. hierarchical softmax.}
  \label{fig:AutoregressiveNetworks}
\end{figure*}

Generative topic models such as LDA or DocNADE infer topic-word distributions that can be used to estimate a document likelihood. 
While basic models such as LDA do not account for context information when inferring these distributions, more recent approaches
such as DocNADE achieve {\it amplified word and document  likelihoods} by accounting for words preceding a word of interest in a document.
More specifically, DocNADE \cite{Hugo:82,Hugo:83} (Figure \ref{fig:AutoregressiveNetworks}, Left) is a probabilistic graphical model that learns topics over sequences of words, corresponding to a language model \cite{Manning:82,Bengio:82} 
that can be interpreted as a neural network with several parallel hidden layers. 
To predict the word $v_i$, each hidden layer ${\bf h}_i$ takes as input the sequence of preceding words ${\bf v}_{<i}$.  
However, it does {\it not} take into account the following words ${\bf v}_{>i}$ in the sequence. 
Inspired by bidirectional language models \cite{Mousa:82} and recurrent neural networks 
\cite{elman1990finding,Gupta:82,Vu:82,Gupta:87}, trained to
predict a word (or label) depending on its full left and right contexts, we extend DocNADE and incorporate 
full contextual information (all words around ${v}_i$) at each hidden layer ${\bf h}_i$ when predicting the word ${v}_i$ in a language modeling fashion with neural topic modeling.


While this is a powerful approach for incorporating contextual information in particular for long texts and corpora with many documents, 
learning contextual information remains challenging in topic models 
with short texts and few documents, 
due to (1) limited word co-occurrences or little context and (2) significant word non-overlap in such short texts. 
However, distributional word  representations (i.e. word embeddings)
have shown to capture both the semantic and syntactic relatedness in words and demonstrated impressive 
performance in natural language processing (NLP) tasks.  
For example, assume that we conduct topic analysis over the two short text fragments: 
``{\it Goldman shares drop sharply downgrade}" and 
``{\it Falling market homes weaken economy}". Traditional topic models will not be able to infer relatedness between word pairs across sentences such as ({\it economy}, {\it shares}) due to the lack of word-overlap between sentences. However, in embedding space, the word pairs ({\it economy}, {\it shares}), ({\it market}, {\it shares}) and  ({\it falling}, {\it drop}) have cosine similarities of $0.65$, $0.56$ and $0.54$.


Therefore, we {\it incorporate word embeddings} as fixed prior in neural topic models 
in order to introduce complementary information.  
The proposed neural architectures learn task specific word vectors in association with static embedding priors leading to better text representation for 
topic extraction, information retrieval, classification, etc. 

The multi-fold {\bf contributions} in this work are: 
{\bf (1)} We propose an advancement in neural autoregressive topic model by incorporating full 
contextual information around words in a document to boost the likelihood of each word (and document). 
This enables learning better ({\it informed}) document representations that we quantify via 
{\it generalization} (perplexity), 
{\it interpretability} (topic coherence)  and 
{\it applicability} (document retrieval and classification). 
We name the proposed topic model as {\it Document Informed Neural Autoregressive Distribution Estimator} ({\bf iDocNADE}).  
{\bf (2)} We propose a further extension of DocNADE-like models by incorporating complementary information via word embeddings, along with the standard sparse word representations 
(e.g., one-hot encoding). The resulting two DocNADE variants are named as 
\textit{Document Neural Autoregressive Distribution Estimator with Embeddings} ({\bf DocNADEe}) and 
\textit{Document Informed Neural Autoregressive Distribution Estimator with Embeddings} ({\bf iDocNADEe}).  
{\bf (3)} We also investigate the two contributions above in the deep versions of topic models.  
{\bf (4)} We apply our modeling approaches to 8 short-text and 7 long-text datasets from diverse domains.  
With the learned representations, we show a gain of $5.2$\% (404 vs 426) in perplexity, $11.1$\% (.60 vs .54)  in precision at retrieval fraction 0.02 and $5.2$\% ($.664$ vs $.631$) in $F1$ for text categorization,
 compared to the DocNADE model (on average over 15 datasets). 
{\it Code} and {\it supplementary material} 
are available at {\tt https://github.com/pgcool/iDocNADEe}.%

\section{Neural Autoregressive Topic Models}
RSM \cite{Salakhutdinov:82}, 
a probabilistic undirected topic model, is a generalization of  the energy-based Restricted Boltzmann
Machines RBM \cite{Hinton:83} 
that can be used to model word counts. 
NADE \cite{Hugo:84} decomposes the joint distribution of observations into autoregressive conditional distributions, 
modeled using non-linear  functions.  
Unlike for RBM/RSM, this leads to tractable 
gradients of the 
data negative 
log-likelihood but can only be used to model binary observations.

{\bf DocNADE} (Figure \ref{fig:AutoregressiveNetworks}, Left) is a generative neural autoregressive topic model to 
account for word counts, inspired by RSM and NADE. 
For a document ${\bf v}=[v_1, ..., v_D]$ of size $D$, it models the joint distribution $p({\bf v})$ of 
all words $v_i$, where $v_i \in \{1,..., K\}$ 
is the index of the $i$th word in the dictionary of vocabulary size $K$. This is achieved by decomposing it as a product of  
conditional distributions i.e. $p({\bf v})= \prod_{i=1}^{D} p(v_i | {\bf v}_{<i})$ 
and computing each autoregressive conditional  $p(v_i | {\bf v}_{<i})$  via a feed-forward neural network for $i \in \{1,...D\}$,\vspace{-0.1cm}   
\begin{eqnarray}
\overrightarrow{\bf h}_i({\bf v}_{<i}) & =  g ({\bf c} + \sum_{k<i} {\bf W}_{:, v_k})\eqlabel{scalingfactor}\\
p (v_i = w | {\bf v}_{<i}) & = \frac{\exp (b_w + {\bf U}_{w,:} \overrightarrow{\bf h}_i ({\bf v}_{<i}))}{\sum_{w'} \exp (b_{w'} + {\bf U}_{w',:} \overrightarrow{\bf h}_i ({\bf v}_{<i}))}\nonumber
\end{eqnarray}
where ${\bf v}_{<i} \in \{v_1, ...,v_{i-1}\}$.  
$g(\cdot)$ is a non-linear activation function, ${\bf W} \in \mathbb{R}^{H \times K}$ and ${\bf U} \in \mathbb{R}^{K \times H}$ 
are weight 
matrices, ${\bf c} \in \mathbb{R}^H$ and ${\bf b} \in \mathbb{R}^K$ are bias parameter 
vectors. $H$ is the number of hidden units (topics). 
${\bf W}_{:,<i}$ is a matrix made of the $i-1$ first columns of ${\bf W}$. 
The probability of the word $v_i$ 
is thus computed using a position-dependent hidden layer $\overrightarrow{\bf h}_i({\bf v}_{<i})$ 
that learns a representation based on all previous words ${\bf v}_{<i}$; 
however it does {\it not} incorporate the following words ${\bf v}_{>i}$. Taken together, the log-likelihood  of any document {\bf v} of arbitrary length can be computed as:\vspace{-0.2cm}
\begin{eqnarray}\label{eq:DocNADEloglikelihood}
\mathcal{L}^{DocNADE}({\bf v})  =  \sum_{i=1}^{D} \log p (v_i | {\bf v}_{<i})
\end{eqnarray}


{\bf iDocNADE} (Figure \ref{fig:AutoregressiveNetworks}, Right), our {\it proposed} model, accounts for the full context information 
(both previous ${\bf v}_{<i}$ and following ${\bf v}_{>i}$ words) around each word $v_i$  for a document ${\bf v}$. 
Therefore, the log-likelihood $\mathcal{L}^{iDocNADE}$  for a document $\bf v$ in {\it iDocNADE} is computed using forward and backward language models as:\vspace{-0.2cm}
\begin{eqnarray} \label{eq:iDocNADEloglikelihood}
\log p({\bf v})  = \frac{1}{2}  \sum_{i=1}^{D}  \underbrace{\log p(v_i | {\bf v}_{<i})}_{\mbox{forward}} & +   \underbrace{\log p(v_i | {\bf v}_{>i})}_{\mbox{backward}}  
\end{eqnarray}
i.e., the mean of the forward ($\overrightarrow{\mathcal{L}}$) and backward ($\overleftarrow{\mathcal{L}}$)  log-likelihoods. 
This is achieved in a bi-directional language modeling 
and feed-forward fashion by computing  position dependent {\it forward} ($\overrightarrow{\bf h}_i$) 
and {\it backward} ($\overleftarrow{\bf h}_i$) hidden layers   
for each word $i$, as:\vspace{-0.1cm} 
\begin{align}
\begin{split}\label{eq:hforward}
\overrightarrow{\bf h}_i({\bf v}_{<i}) =  g (\overrightarrow{\bf c} +\sum_{k<i} {\bf W}_{:, v_k}) 
\end{split}\\
\begin{split}\label{eq:hbackward}
\overleftarrow{\bf h}_i({\bf v}_{>i}) =  g (\overleftarrow{\bf c} + \sum_{k>i} {\bf W}_{:, v_k}) 
\end{split}
\end{align}
where $\overrightarrow{\bf c} \in \mathbb{R}^H$ and $\overleftarrow{\bf c} \in \mathbb{R}^H$ are bias parameters 
in forward and backward passes, respectively. $H$ is the number of hidden units (topics). 

Two autoregressive conditionals are computed for each $i$th word using the forward and backward hidden vectors,
\vspace{-0.1cm}  
\begin{align}
\begin{split}\label{eq:pvforward}
p (v_i = w | {\bf v}_{<i}) = \frac{\exp (\overrightarrow{b}_w + {\bf U}_{w,:} \overrightarrow{\bf h}_i ({\bf v}_{<i}))}{\sum_{w'} \exp (\overrightarrow{b}_{w'} + {\bf U}_{w',:} \overrightarrow{\bf h}_i ({\bf v}_{<i}))}
\end{split}\\
\begin{split}\label{eq:pvbackward}
p (v_i = w | {\bf v}_{>i}) = \frac{\exp (\overleftarrow{b}_w + {\bf U}_{w,:} \overleftarrow{\bf h}_i ({\bf v}_{>i}))}{\sum_{w'} \exp (\overleftarrow{b}_{w'} + {\bf U}_{w',:} \overleftarrow{\bf h}_i ({\bf v}_{>i}))}
\end{split}
\end{align}
for  $i \in [1, ..., D]$ where  $\overrightarrow{\bf b} \in \mathbb{R}^K$ and $\overleftarrow{\bf b} \in \mathbb{R}^K$ are biases in forward and 
backward passes, respectively.  Note that the parameters {\bf W} and {\bf U} are shared between the two networks. 
\begin{algorithm}[t]
\caption{{\small Computation of $\log p({\bf v})$ in {\it iDocNADE} or {\it iDocNADEe} 
using {\it tree-softmax} or {\it full-softmax}}}\label{trainingiDocNADE}
\small{ 
\begin{algorithmic}[1]
\Statex \textbf{Input}: A training document vector {\bf v},  Embedding matrix {\bf E} 
\Statex \textbf{Parameters}: \{\overrightarrow{\bf b},  \overleftarrow{\bf b}, \overrightarrow{\bf c}, \overleftarrow{\bf c}, {\bf W}, {\bf U}\}
\Statex \textbf{Output}: $\log p({\bf v})$
\State $\overrightarrow{\bf a} \gets \overrightarrow{\bf c}$  
\If {\texttt{iDocNADE}} \State  $\overleftarrow{\bf a} \gets \overleftarrow{\bf c} +  \sum_{i >1}{\bf W}_{:, v_{i}}$ \EndIf
\If {\texttt {iDocNADEe}} \State $\overleftarrow{\bf a} \gets \overleftarrow{\bf c} +  \sum_{i >1}{\bf W}_{:, v_{i}} + \lambda \sum_{i >1}{\bf E}_{:, v_{i}}$ \EndIf
\State $ q({\bf v}) = 1$
\For{$i$ from $1$ to $D$}
        \State $\overrightarrow{\bf h}_{i}  \gets g(\overrightarrow{\bf a})$; \ \  $\overleftarrow{\bf h}_{i}  \gets g(\overleftarrow{\bf a})$
        \If {\texttt{tree-softmax}}     
        \State $  p(v_{i} | {\bf v}_{<i}) = 1$; \ \   $  p(v_{i} | {\bf v}_{>i}) = 1$
        \For{$m$ from $1$ to $| {\bf \pi}(v_{i})|$}
                   \State $  p(v_{i} | {\bf v}_{<i}) \gets   p(v_{i} | {\bf v}_{<i})    p({\bf \pi}(v_{i})_m| {\bf v}_{<i})$
                   \State $  p(v_{i} | {\bf v}_{>i}) \gets   p(v_{i} | {\bf v}_{>i})    p({\bf \pi}(v_{i})_m| {\bf v}_{>i})$
        \EndFor
       \EndIf
       
       \If {\texttt{full-softmax}} 
            \State compute $p(v_{i} | {\bf v}_{<i})$ using equation \ref{eq:pvforward} 
             \State compute $p(v_{i} | {\bf v}_{>i}) $ using equation \ref{eq:pvbackward} 
        \EndIf

	\State $ q({\bf v}) \gets  q({\bf v})   p(v_{i} | {\bf v}_{<i})  p(v_{i} | {\bf v}_{>i})$

       \If {\texttt{iDocNADE}} \State $\overrightarrow{\bf a} \gets \overrightarrow{\bf a} + {\bf W}_{:, v_{i}}$ ; \ \  $\overleftarrow{\bf a} \gets \overleftarrow{\bf a} - {\bf W}_{:, v_{i}}$ 
      \EndIf
      \If  {\texttt{iDocNADEe}} \State $\overrightarrow{\bf a} \gets \overrightarrow{\bf a} + {\bf W}_{:, v_{i}} + \lambda \ {\bf E}_{:, v_{i}}$ 
     \State  $\overleftarrow{\bf a} \gets \overleftarrow{\bf a} - {\bf W}_{:, v_{i}} - \lambda \ {\bf E}_{:, v_{i}}$ 
     \EndIf
\EndFor
\State $\log p({\bf v}) \gets \frac{1}{2} \log q({\bf v})$
\end{algorithmic}}
\end{algorithm}

{\bf DocNADEe and iDocNADEe with Embedding priors}: 
We introduce additional semantic information for each word into DocNADE-like models via its pre-trained embedding vector, thereby enabling better textual representations and semantically more coherent topic distributions, in particular for short texts.   
In its simplest form, we extend DocNADE with word embedding aggregation at each autoregressive step $k$ 
to generate a complementary textual representation, i.e., $\sum_{k<i} {\bf E}_{:, v_k}$. 
This mechanism utilizes prior knowledge encoded in a pre-trained embedding matrix  {${\bf E} \in  \mathbb{R}^{H \times K}$} when learning 
task-specific matrices {$ \bf W$} and latent representations in DocNADE-like models.   
The position dependent forward $\overrightarrow{{\bf h}_i^{e}}({\bf v}_{<i})$ 
and (only in iDocNADEe)  backward $\overleftarrow{{\bf h}_i^{e}}({\bf v}_{>i})$ hidden layers for each word $i$ now depend on $\bf E$ as:
\begin{align}
\begin{split}
\overrightarrow{{\bf h}_i^{e}}({\bf v}_{<i}) =  g (\overrightarrow{\bf c} +\sum_{k<i} {\bf W}_{:, v_k} + \lambda \sum_{k<i} {\bf E}_{:, v_k}) 
\end{split}\\
\begin{split}
\overleftarrow{{\bf h}_i^{e}}({\bf v}_{>i}) =  g (\overleftarrow{\bf c} + \sum_{k>i} {\bf W}_{:, v_k} + \lambda \sum_{k>i} {\bf E}_{:, v_k}) 
\end{split}
\end{align}

where, $\lambda$ is a mixture coefficient, determined using validation set. As in  equations \ref{eq:pvforward} and \ref{eq:pvbackward}, 
the forward and backward autoregressive conditionals are computed via hidden vectors $\overrightarrow{{\bf h}_i^{e}}({\bf v}_{<i})$ and $\overleftarrow{{\bf h}_i^{e}}({\bf v}_{>i})$, respectively.  

{\bf Deep DocNADEs with/without Embedding Priors}:
DocNADE can be extended to a deep, multiple hidden layer architecture by adding new hidden layers as in a regular deep feed-forward neural network, allowing for improved performance \cite{HugoJMLR:82}. 
In this deep version of DocNADE variants, the first hidden layers are computed in an analogous fashion to iDocNADE (eq. \ref{eq:hforward} and \ref{eq:hbackward}). Subsequent hidden layers are computed as:\vspace{-0.1cm}
\[
\overrightarrow{{\bf h}_i}^{(d)}({\bf v}_{<i}) = g (\overrightarrow{{\bf c}}^{(d)} + {\bf W}^{(d)} \cdot \overrightarrow{{\bf h}_i}^{(d-1)}({\bf v}_{<i}))
\]
and similarly, $\overleftarrow{{\bf h}_i}^{(d)}({\bf v}_{>i})$ 
for $d=2,...,n$, where $n$ is the total number of hidden
layers. The exponent ``$(d)$'' is used as an index over the
hidden layers and parameters in the deep feed-forward
network.  
Forward and/or backward conditionals
for each word $i$ 
are modeled using the 
forward and backward hidden vectors at
the last layer $n$.  The deep DocNADE or iDocNADE variants
without or with embeddings are named as {\it DeepDNE}, {\it
iDeepDNE}, {\it DeepDNEe} and {\it
iDeepDNEe}, respectively 
where ${\bf W}^{(1)}$ is the word representation matrix. 
However in {\it DeepDNEe} (or {\it iDeepDNEe}), we introduce embedding prior $\bf E$ in the first hidden layer, 
i.e., \[ \overrightarrow{{\bf h}_i}^{e, (1)} = g (\overrightarrow{{\bf c}}^{(1)} +\sum_{k<i} {\bf W}_{:, v_k}^{(1)} + \lambda \sum_{k<i} {\bf E}_{:, v_k})\] 
for each word $i$ via embedding aggregation of its context ${\bf v}_{<i}$ (and ${\bf v}_{>i}$). Similarly, we compute $\overleftarrow{{\bf h}_i}^{e, (1)}$. 

{\bf Learning}: 
Similar to DocNADE, the 
conditionals  $p (v_i = w | {\bf v}_{<i})$ and $p (v_i = w | {\bf v}_{>i})$ 
in DocNADEe, iDocNADE or iDocNADEe are computed by a neural network for each word $v_i$, allowing efficient learning of {\it informed} 
representations $\overrightarrow{\bf h}_{i}$ and  $\overleftarrow{\bf h}_{i}$ (or 
$\overrightarrow{{\bf h}_i^{e}}({\bf v}_{<i})$ and $\overleftarrow{{\bf h}_i^{e}}({\bf v}_{>i})$), 
as it consists simply of a linear transformation followed by a  non-linearity. 
Observe that the weight $\bf W$ (or prior embedding matrix $\bf E$) is the same across all conditionals and ties contextual observables (blue colored lines in Figure 1)  by computing each $\overrightarrow{\bf h}_{i}$ 
or $\overleftarrow{\bf h}_{i}$ 
(or $\overrightarrow{{\bf h}_i^{e}}({\bf v}_{<i})$ and $\overleftarrow{{\bf h}_i^{e}}({\bf v}_{>i})$).   

{\it Binary word tree (\texttt{tree-softmax}) to compute conditionals}: 
To compute the likelihood of a document, the autoregressive conditionals  $p (v_i = w | {\bf v}_{<i})$ and $p (v_i = w | {\bf v}_{>i})$ have to be computed for each word $i \in [1, 2, ...D]$, requiring time linear in vocabulary size $K$.
To reduce computational cost and achieve a complexity logarithmic in $K$ we follow \citeauthor{Hugo:82} \shortcite{Hugo:82} and decompose the computation of the conditionals using a probabilistic tree. All words in the documents are randomly assigned to a different leaf in a binary tree 
and the probability of a word is computed as the probability of reaching its associated leaf from the root. 
Each left/right transition probability is modeled using a binary logistic regressor  with 
the hidden layer $\overrightarrow{\bf h}_{i}$ or  $\overleftarrow{\bf h}_{i}$ ($\overrightarrow{{\bf h}_{i}^{e}}$ or  $\overleftarrow{{\bf h}_{i}^{e}}$) as its input.  
In the binary tree, the probability of a given word is computed by multiplying each of the left/right 
transition probabilities  along the tree path.  

Algorithm \ref{trainingiDocNADE} shows the computation of $\log p({\bf v})$ using {\it iDocNADE} (or {\it iDocNADEe}) structure, where the autogressive conditionals (lines 14 and 15) for each word $v_i$ are obtained from the forward and backward networks and modeled into a binary word tree, where $\pi (v_{i})$ denotes the sequence of 
binary left/right choices at the internal nodes along the tree path and ${\bf l}(v_i)$ the sequence of tree nodes on that tree path. 
For instance, $l (v_i)_1$ will always be the root of the binary tree and $\pi (v_i)_1$ will be 0 if the word leaf $v_i$ is in the left subtree or 1 otherwise.   
Therefore, each of the forward and backward conditionals are computed as: \vspace{-0.3cm}

{\small 
\begin{align*}
\begin{split}
p(v_i = w | {\bf v}_{<i}) & = \prod_{m=1}^{|\pi (v_i)|} p(\pi (v_i)_m | {\bf v}_{<i})
\end{split}\\
\begin{split}
p(v_i = w | {\bf v}_{>i}) & = \prod_{m=1}^{|\pi (v_i)|} p(\pi (v_i)_m | {\bf v}_{>i})
\end{split}\\
\begin{split}
p(\pi (v_i)_m | {\bf v}_{<i}) = & g( \overrightarrow{b}_{l{(v_i)_m}} + {\bf U}_{l{(v_i)_m}, :}  \overrightarrow{\bf h}({\bf v}_{<i})) 
\end{split}\\
\begin{split}
p(\pi (v_i)_m | {\bf v}_{>i})  = & g( \overleftarrow{b}_{l{(v_i)_m}} + {\bf U}_{l{(v_i)_m}, :}  \overleftarrow{\bf h}({\bf v}_{>i})) 
\end{split}
\end{align*}}

where  ${\bf U} \in \mathbb{R}^{T \times H}$ is the matrix of logistic regressions weights, $T$ is the number of internal nodes in binary tree, 
and  \overrightarrow{\bf b} and \overleftarrow{\bf b} are bias vectors. 

\enote{hs}{are you sure that you have introduced $T$:
``where'' usually means: now I'm telling you what these
variables are, but it's not used for variables that have not
been mentioned}


Each of the forward and backward conditionals $p (v_i = w |
{\bf v}_{<i})$ or $p (v_i = w | {\bf v}_{>i} )$ requires the
computation of its own hidden layers $\overrightarrow{\bf
h}_i ({\bf v}_{<i})$ and $\overleftarrow{\bf h}_i ({\bf
v}_{>i})$ (or $\overrightarrow{{\bf h}_i^{e}} ({\bf
v}_{<i})$ and $\overleftarrow{{\bf h}_i^{e}} ({\bf
v}_{>i})$), respectively.  With $H$ being the size of each
hidden layer and $D$ the number of words in ${\bf v}$,
computing a single layer requires $O(HD)$, and since there
are $D$ hidden layers to compute, a naive approach for
computing all hidden layers would be in $O(D^2H)$. However,
since the weights in the matrix $\bf W$ are tied, the
linear activations $\overrightarrow{\bf a}$ and
$\overleftarrow{\bf a}$ (algorithm \ref{trainingiDocNADE})
can be re-used in every hidden layer and computational
complexity reduces to $O(HD)$.


With the trained {\it iDocNADEe} (or {\it DocNADE} variants), the representation ($\overleftrightarrow{{\bf \mathfrak{h}}^{e}} \in { \mathbb R}^{H}$) for a new document {\bf v}*  of size $D^*$ is extracted by summing 
the hidden representations from the forward and backward networks to account for the  
context information around each word in the words' sequence, as\vspace{0.3cm} 
{\small \begin{align}
\begin{split}
\overrightarrow{{\bf h}^{e}} ({\bf v}^*)  =  g
&(\overrightarrow{\bf c} + \sum_{k\leq D^*} {\bf W}_{:,v_k^*} + \lambda \sum_{k\leq D^*} {\bf E}_{:, v_k^*}) 
\end{split}\\
\begin{split}
\overleftarrow{{\bf h}^{e}}({\bf v}^*)  =  g
&(\overleftarrow{\bf c} +\sum_{k\geq 1} {\bf W}_{:,v_k^*} + \lambda \sum_{k\geq 1} {\bf E}_{:, v_k^*})
\end{split}\\
\label{eq:documentrepiDocNADE}
\begin{split}
\mbox{Therefore;} &\  \overleftrightarrow{{\bf \mathfrak{h}}^{e}}  = \overrightarrow{{\bf h}^{e}}(\bf v^*) + \overleftarrow{{\bf h}^{e}}(\bf v^*)
\end{split}   
\end{align}}
The DocNADE variants without embeddings compute the representation $\overleftrightarrow{\bf \mathfrak{h}}$ excluding the   embedding term ${\bf E}$. 
Parameters \{\overrightarrow{\bf b},  \overleftarrow{\bf b}, \overrightarrow{\bf c}, \overleftarrow{\bf c}, {\bf W}, {\bf U}\} are 
learned by minimizing the average negative log-likelihood of the training documents using stochastic gradient descent (algorithm \ref{gradientiDocNADE}). 
In our proposed formulation of iDocNADE or its variants (Figure \ref{fig:AutoregressiveNetworks}), we perform 
inference by computing $\mathcal{L}^{iDocNADE}({\bf v})$ (Eq.\ref{eq:iDocNADEloglikelihood}). 


\begin{algorithm}[t]
\caption{{\small Computing gradients of $-\log p({\bf v})$ in {\it iDocNADE} or {\it iDocNADEe} using 
{\it tree-softmax}} 
}\label{gradientiDocNADE}
\small{ 
\begin{algorithmic}[1]
\Statex \textbf{Input}: A training document vector {\bf v}
\Statex \textbf{Parameters}: \{\overrightarrow{\bf b},  \overleftarrow{\bf b}, \overrightarrow{\bf c}, \overleftarrow{\bf c}, {\bf W}, {\bf U}\}
\Statex \textbf{Output}: $\delta\overrightarrow{\bf b},  \delta\overleftarrow{\bf b}, \delta\overrightarrow{\bf c}, \delta\overleftarrow{\bf c}, \delta{\bf W}, \delta{\bf U}$ 
\State $\overrightarrow{\bf a} \gets 0$; $\overleftarrow{\bf a} \gets 0$; $\overrightarrow{\bf c} \gets 0$;  $\overleftarrow{\bf c} \gets 0$; $\overrightarrow{\bf b} \gets 0$;  $\overleftarrow{\bf b} \gets 0$ 
\For{$i$ from $D$ to $1$}
        \State $\delta\overrightarrow{\bf h}_i \gets 0$ ; \ \ $\delta\overleftarrow{\bf h}_i \gets 0$
        \For{$m$ from $1$ to $| {\bf \pi}(v_{i})|$}
               \State $\overrightarrow{b}_{l{(v_i)_m}} \gets \overrightarrow{b}_{l{(v_i)_m}} + ( p(\pi(v_i)_m | {\bf v}_{<i}) - \pi(v_i)_m)$
               \State $\overleftarrow{b}_{l{(v_i)_m}} \gets \overleftarrow{b}_{l{(v_i)_m}} + ( p(\pi(v_i)_m | {\bf v}_{>i}) - \pi(v_i)_m)$
               \State $\delta\overrightarrow{\bf h}_i \gets \delta\overrightarrow{\bf h}_i  + (p( \pi(v_i)_m | {\bf v}_{<i}) -  \pi(v_i)_m) {\bf U}_{l(v_i)m,:}$               
               \State $\delta\overleftarrow{\bf h}_i \gets \delta\overleftarrow{\bf h}_i  + (p( \pi(v_i)_m | {\bf v}_{>i}) -  \pi(v_i)_m) {\bf U}_{l(v_i)m,:}$
              \State $\delta{\bf U}_{l(v_i)_m}  \gets \delta{\bf U}_{l(v_i)_m} + ( p(\pi(v_i)_m | {\bf v}_{<i}) - \pi(v_i)_m) \overrightarrow{\bf h}_i^{T} +  (p(\pi(v_i)_m | {\bf v}_{>i}) - \pi(v_i)_m) \overleftarrow{\bf h}_i^{T}$
        \EndFor
        \State $\delta\overrightarrow{\bf g} \gets \overrightarrow{\bf h}_i  \circ (1 - \overrightarrow{\bf h}_i)$ \# for sigmoid activation
        \State $\delta\overleftarrow{\bf g} \gets \overleftarrow{\bf h}_i  \circ (1 - \overleftarrow{\bf h}_i)$ \# for sigmoid  activation
        \State $\delta\overrightarrow{\bf c} \gets  \delta\overrightarrow{\bf c} + \delta\overrightarrow{\bf h}_i  \circ \delta\overrightarrow{\bf g}$ \ ; \ $\delta\overleftarrow{\bf c} \gets  \delta\overleftarrow{\bf c} + \delta\overleftarrow{\bf h}_i  \circ \delta\overleftarrow{\bf g}$
        \State $ \delta{\bf W}_{:,v_i} \gets  \delta{\bf W}_{:,v_i} + \delta\overrightarrow{\bf a} +  \delta\overleftarrow{\bf a}$
        \State $\delta\overrightarrow{\bf a} \gets \delta\overrightarrow{\bf a} +  \delta\overrightarrow{\bf h}_i  \circ \delta\overrightarrow{\bf g}$ \ ; \  $\delta\overleftarrow{\bf a} \gets \delta\overleftarrow{\bf a} +  \delta\overleftarrow{\bf h}_i  \circ \delta\overleftarrow{\bf g}$
\EndFor
\end{algorithmic}} 
\end{algorithm}

\begin{table*}[t]
\center
\small
\renewcommand*{\arraystretch}{1.2}
\resizebox{.99\textwidth}{!}{
\begin{tabular}{r|rrrrrrr||rr|rr||rr|rr|rr|rr}

 \multirow{3}{*}{\bf Data} &  \multirow{3}{*}{\bf Train} &  \multirow{3}{*}{\bf Val} & \multirow{3}{*}{\bf Test} & \multirow{3}{*}{\bf K} & \multirow{3}{*}{\bf L} & \multirow{3}{*}{\bf C} & \multirow{3}{*}{\bf Domain}      & \multicolumn{4}{c||}{\texttt{Tree-Softmax(TS)}}    & \multicolumn{8}{c}{\texttt{Full-Softmax (FS)}}\\ \cline{9-20}
        
  &   & &  & & & & & \multicolumn{2}{c|}{{\bf DocNADE}}  & \multicolumn{2}{c||}{\bf iDocNADE}     & \multicolumn{2}{c|}{\bf DocNADE}    & \multicolumn{2}{c|}{\bf iDocNADE}     & \multicolumn{2}{c|}{\bf DocNADEe}    & \multicolumn{2}{c}{\bf iDocNADEe}  \\   \cline{9-20}

  &   & &  & & & & & PPL   &  IR       &  PPL   &  IR     &  PPL   &  IR     &  PPL   &  IR     &  PPL   &  IR   &  PPL   &  IR   \\ \hline

20NSshort             & 1.3k & 0.1k & 0.5k &    2k       &  13.5    &   20      &     News            &   894      &    .23                &     880     &     .30           &    646       &  .25        &      639     &   .26        &     638      &    .28               &    \underline{633}        &    {\bf .28}       \\

TREC6                  & 5.5k  &  0.5k &  0.5k  &      2k          &   9.8   & 6        &   Q\&A       &    42      &    .48               &     39     &     .55              &   64       &  .54        &      61     &   .56        &      62      &    .56             &   \underline{60}        &      {\bf .57}         \\

R21578title$^\dagger$ &  7.3k &  0.5k & 3.0k     &   2k          &   7.3   &     90  &   News    &    298      &    .61               &     239     &     .63           &   193       &  .61       &      181     &   .62      &     179      &    .65             &    \underline{176}        &      {\bf .66}         \\

Subjectivity           &  8.0k &  .05k &  2.0k  & 2k       &   23.1   &  2                &   Senti      &    303      &    .78                &     287     &     .81           &  371       &  .77          &     365     &   .80        &      362      &    .80              &   \underline{361}        &      {\bf .81}         \\

Polarity                 & 8.5k &  .05k & 2.1k &    2k       &   21.0   &   2              &  Senti      &    311      &    .51                &     292     &     .54            &   358       &  .54          &   345     &   .56          &     341      &    .56             &    \underline{340}        &      {\bf .57}       \\

TMNtitle                & 22.8k  &  2.0k & 7.8k  &       2k        &   4.9   &     7    &   News    &    863      &    .57                &     823     &     .59            &   711       &  .44         &  670     &   .46        &      668      &    .54               &   \underline{664}        &      {\bf .55}         \\

TMN                 & 22.8k  &  2.0k & 7.8k  &      2k        &    19  &       7   &  News            &    548      &    .64                &     536     &     .66               &   592       &  .60         &  560     &   .64         &     563      &    .64            &    \underline{561}        &     {\bf  .66}    \\

AGnewstitle           & 118k &  2.0k &  7.6k  &        5k         &  6.8    &    4      &  News   &    811      &    .59               &     793     &     .65           &   545       &  .62          &  516     &   .64        &     516      &    .66              &    \underline{514}        &      {\bf .68}       \\ \hline


 \multicolumn{1}{c}{\bf Avg (short)}    &      &     &     &              &      &           &   &    509        &    .55                &     486     &     .59           &   435       &  .54          &   417     &   .57        &     416      &    .58               &    \underline{413}        &     {\bf  .60}  \\ \hline \hline

20NSsmall             & 0.4k &  0.2k & 0.2k  &      2k       &   187   &    20      & News            &    -      &    -                &     -     &     -             &   628       &  .30         &  592     &   .32         &     607      &    {\bf  .33}             &    \underline{590}        &     {\bf  .33}       \\
Reuters8             & 5.0k &  0.5k & 2.2k  &      2k       &   102   &    8      & News            &    172      &    .88                &     152     &     .89             &   184       &  .83         &  \underline{178}     &   .88         &     \underline{178}      &     {\bf  .87}             &    \underline{178}        &     {\bf  .87}       \\

20NS                  & 8.9k & 2.2k & 7.4k  &     2k       &   229   &   20       &   News          &    830      &    .27               &     812     &     .33               &   474       &  .20         &  \underline{463}        &   .24         &      464      &    {\bf .25}          &   \underline{463}        &      {\bf .25}     \\

R21578$^\dagger$ &  7.3k&  0.5k & 3.0k  &      2k        &   128   &   90      &    News       &    215      &    .70                &     179     &     .74               &   297       &  .70          &   \underline{285}     &   {\bf .73}        &      286      &    .71              &    \underline{285}        &      .72    \\

RCV1V2$^\dagger$ &  23.0k   &  .05k    &    10.0k        &    2k   &  123        &     103      &  News      &    381      &    .81              &    364      &     .86                &   479       &  .86          &  463     &   {\bf .89}        &      465      &    .87          &   \underline{462}        &    .88     \\ 

SiROBs$^\dagger$ &  27.0k &  1.0k & 10.5k  &      3k       &  39    &  22    &  Industry   &    398      &    .31                &     351     &     .35                &   399       &  .34         &  \underline{340}     &   .34        &     343      &    {\bf .37}             &   \underline{340}        &      .36   \\

AGNews              & 118k & 2.0k &   7.6k &        5k        &  38    &     4    &   News      &    471      &    .72              &     441     &     .77                 &   451       &  .71       &   439     &   .78        &      \underline{433}      &    .76           &   438        &      {\bf .79}   \\

\hline

 \multicolumn{1}{c}{\bf Avg (long)}     &      &     &     &              &      &           &     &    417      &    .61           &     383     &     .65               &   416       &  .56          &   394     &   {\bf  .60}       &     396       &    {\bf  .60}           &   \underline{393}        &     {\bf  .60}     \\ \hline \hline 

 \multicolumn{1}{c}{\bf Avg (all)}      &      &     &     &              &      &           &        &    469       &    .57            &     442     &     .62                &   426       &  .54           &  406     &   .58      &     407      &    .59             &    \underline{404}        &     {\bf  .60}   

\end{tabular}}
\caption{{\it Data statistics} of short and long texts as well as small and large corpora from various domains. {\it State-of-the-art} comparison in terms of PPL and IR (i.e, IR-precision) for {\bf short} and {\bf long} text datasets. 
The symbols are- $L$: average text length in number of words, $K$:dictionary size, $C$: number of classes, Senti: Sentiment, Avg: average, `k':thousand and $\dagger$: multi-label data.   
PPL and IR (IR-precision) are computed  over 200 ($T200$)  topics at retrieval fraction = 0.02.  For short-text,  $L < 25$. 
The \underline{\it underline} and {\bf bold} numbers indicate the best scores in PPL and retrieval task, respectively in FS setting. 
See \citeauthor{Hugo:82} \shortcite{Hugo:82} for LDA \cite{Blei:81} performance in terms of PPL, where DocNADE outperforms LDA.}
\label{datastatisticsPPLIRshortlongtext}
\end{table*}

\section{Evaluation}
We perform 
evaluations on 15 (8 short-text and 7 long-text) datasets 
of varying size with single/multi-class labeled documents from public as well as industrial corpora. See the {\it supplementary material} for the data description, hyper-parameters and grid-search results for generalization and IR tasks. 
Table \ref{datastatisticsPPLIRshortlongtext} shows the data statistics, where 20NS: 20NewsGroups and R21578: Reuters21578.   
Since, \citeauthor{GuptatextTOvec:81} \shortcite{GuptatextTOvec:81} have shown that DocNADE outperforms gaussian-LDA \cite{Das:82}, glove-LDA and glove-DMM \cite{nguyen2015gloveldaglovegmm} in terms of topic coherence, text retrieval and classification, therefore 
we adopt DocNADE as the strong {\it baseline}.    
We use the development (dev) sets of each of the datasets 
to perform a grid-search on mixture weights, $\lambda = [0.1, 0.5, 1.0]$. 

\makeatletter
\def\labelonly{BDF}
\def\labelcheck#1{
    \edef\pgfmarshal{\noexpand\pgfutil@in@{#1}{\labelonly}}
    \pgfmarshal
    \ifpgfutil@in@[#1]\fi
}
\makeatother

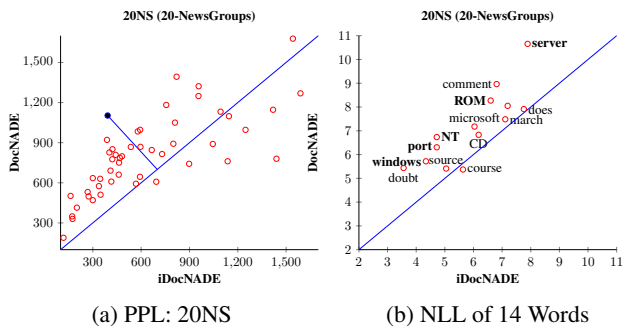
\begin{figure}[t]
\centering
\begin{subfigure}{0.24\textwidth}
\centering
\begin{tikzpicture}[scale=0.5, trim axis right][baseline]
\begin{axis}[%
axis x line=bottom,
axis y line=left,
axis line style={-},
title={\bf 20NS (20-NewsGroups)},
xmin=100,
xmax=1700,
ymin=100, 
ymax=1700,
/pgfplots/xtick={300,600,...,1700},
/pgfplots/ytick={300,600,...,1700},
xlabel= {\bf iDocNADE},
ylabel= {\bf DocNADE},
scatter/classes={%
    a={mark=o,draw=red}, b={draw=blue}}]
\addplot[scatter,only marks,%
    scatter src=explicit symbolic]%
table[meta=label] {
x y label
415 609 a
421 776 a
596 869 a
162  502 a
118  189 a
820 1393 a
403  827 a
1045 890 a
410 691 a
593 645 a
1542 1678 a
957 1322 a
201 414 a
346 630 a
536 868 a
481 798 a
1248 996  a
810 1050 a 
468 786 a
666 845 a
174 330 a
756  1182 a
594 996 a 
1093 1132 a
338 575 a
956 1249 a
2380 1385 a
579  984 a
1589 1269 a
274 498 a
1419 1146 a 
172  349 a
268  531 a 
463  751 a 
568  593 a 
388  921 a
1145  1098 a
300 635 a
392 1103 b
800  892 a
694 608 a
461 661 a
1440 780 a
300 470 a
348 511 a
442 809 a
422 851 a
730 815 a
1137 761 a
898 742 a
    };


\addplot[color=blue] coordinates {
	(100,100) (1700,1700)
};
\addplot[color=blue] coordinates {
	(392,1103) (700,700)
};
\end{axis}
\end{tikzpicture}%
\caption{PPL: 20NS} \label{PPL: 20NewsGroup}
\end{subfigure}\hspace*{\fill}%
~%
\begin{subfigure}{0.24\textwidth}
\centering
\begin{tikzpicture}[scale=0.5,trim axis right,trim axis left][baseline]
\begin{axis}[%
axis x line=bottom,
axis y line=left,
axis line style={-},
title={\bf 20NS (20-NewsGroups)},
xmin=2.0,
xmax=11.0,
ymin=2.0, 
ymax=11.0,
/pgfplots/xtick={2.0,...,11.0},
/pgfplots/ytick={2.0,...,11.0},
xlabel= {\bf iDocNADE},
ylabel= {\bf DocNADE},,
scatter/classes={%
    a={mark=o,draw=red}}]
\addplot[scatter,only marks,%
    scatter src=explicit symbolic]%
table[meta=label] {
x y label
6.18075456 6.83192918 a
6.80839485 8.96412021 a
5.63295427 5.37235833 a
7.75847918 7.91195391 a
3.5600748 5.43678675 a
7.1103252 7.48475653 a
6.0352862 7.17878941 a
4.71765834 6.73556674 a
7.19305586 8.04897594 a
4.71765834 6.30657145 a
6.60034615 8.27238095 a
7.88924478 10.65654915 a
5.04651563 5.41208745 a
4.33936622 5.71744183 a
    };

\node [below] at (axis cs: 6.18075456,6.83192918) {CD};
\node [left] at (axis cs: 6.80839485, 8.96412021) {comment}; 
\node [right] at (axis cs: 5.63295427, 5.37235833) {course}; 
\node [right] at (axis cs: 7.75847918 ,7.91195391) {does}; 
\node [below] at (axis cs: 3.5600748 ,5.43678675 ) {doubt};
\node [right] at (axis cs: 7.1103252, 7.48475653 ) {march};
\node [above] at (axis cs: 6.0352862, 7.17878941) {microsoft}; 
\node [right] at (axis cs: 4.71765834, 6.73556674) {\bf NT}; 
\node [left] at (axis cs: 4.71765834, 6.30657145) {\bf port}; 
\node [left] at (axis cs: 6.60034615, 8.27238095) {\bf ROM}; 
\node [right] at (axis cs: 7.88924478, 10.6565491) {\bf server};
\node [above] at (axis cs: 5.04651563, 5.41208745) {source}; 
\node [left] at (axis cs: 4.33936622, 5.71744183) {\bf windows};

\addplot[color=blue] coordinates {
	(2.0,2.0) (11.0,11.0)
};
\end{axis}
\end{tikzpicture}%
\caption{NLL of 14 Words}\label{NLL20newsgroup}
\end{subfigure}
\caption{{(a)} PPL (T200) by iDocNADE and DocNADE for each of the 50 held-out documents of 20NS.   
The {\it filled circle} 
points to the document for which {\it PPL} differs by maximum.   
{(b)} NLL of each of the words in the document marked by the {\it filled circle} in (a),   
due to iDocNADE and DocNADE.} 
\label{fig:PLandIR}
\end{figure}
\makeatletter

\subsubsection{Generalization (Perplexity, PPL)}
We evaluate the topic models' generative performance as a generative model  of documents 
by estimating log-probability for the test documents. 
During training, we initialize the proposed DocNADE extensions with DocNADE, i.e., {\bf W} matrix. 
A comparison is made with the {\it baselines} (DocNADE and DeepDNE) and proposed variants (iDocNADE, DocNADEe, iDocNADEe, 
iDeepDNE, DeepDNEe and iDeepDNEe)     
using 50 (in {\it supplementary}) and 200 (T200) topics, set by the hidden layer size $H$.  

\makeatletter
\def\labelonly{BDF}
\def\labelcheck#1{
    \edef\pgfmarshal{\noexpand\pgfutil@in@{#1}{\labelonly}}
    \pgfmarshal
    \ifpgfutil@in@[#1]\fi
}
\makeatother

\begin{figure*}[t]
\centering

\begin{subfigure}{0.30\textwidth}
\centering
\begin{tikzpicture}[scale=0.65][baseline]
\begin{axis}[
    xlabel={\bf Fraction of Retrieved Documents (Recall)},
    ylabel={\bf Precision (\%)},
    xmin=0, xmax=9,
    ymin=0.29, ymax=0.72,
   /pgfplots/ytick={.30,.35,...,.72},
    xtick={0,1,2,3,4,5,6,7,8,9},
    xticklabels={0.0001, 0.0005, 0.001, 0.002, 0.005, 0.01, 0.02, 0.05, 0.1, 0.2},
    x tick label style={rotate=45,anchor=east},
    legend pos=south west,
    ymajorgrids=true, xmajorgrids=true,
    grid style=dashed,
]

\addplot[
	color=cyan,
	mark=triangle,
	]
	plot coordinates {
    (0, 0.65)
    (1, 0.61)
    (2, 0.60)
    (3, 0.58)
    (4, 0.56)
    (5, 0.55)
    (6, 0.53)
    (7, 0.49)
    (8, 0.45)
    (9, 0.38)
	};
\addlegendentry{iDeepDNEe}

\addplot[
	color=orange,
	mark=*,
	]
	plot coordinates {
    (0, 0.60)
    (1, 0.55)
    (2, 0.52)
    (3, 0.50)
    (4, 0.45)
    (5, 0.42)
    (6, 0.39)
    (7, 0.35)
    (8, 0.32)
    (9, 0.28)
	};
\addlegendentry{DeepDNE}

\addplot[
	color=blue,
	mark=square,
	]
	plot coordinates {
    (0, 0.69)
    (1, 0.65)
    (2, 0.63)
    (3, 0.62)
    (4, 0.59)
    (5, 0.57)
    (6, 0.54)
    (7, 0.50)
    (8, 0.45)
    (9, 0.38)
	};
\addlegendentry{iDocNADEe}

\addplot[
	color=red,
	mark=square,
	]
	plot coordinates {
    (0, 0.64)
    (1, 0.59)
    (2, 0.57)
    (3, 0.55)
    (4, 0.52)
    (5, 0.49)
    (6, 0.46)
    (7, 0.42)
    (8, 0.38)
    (9, 0.33)
	};
\addlegendentry{iDocNADE}

\addplot[
	color=violet,
	mark=square,
	]
	plot coordinates {
    (0, 0.69)
    (1, 0.65)
    (2, 0.63)
    (3, 0.61)
    (4, 0.59)
    (5, 0.56)
    (6, 0.54)
    (7, 0.49)
    (8, 0.45)
    (9, 0.38)
	};
\addlegendentry{DocNADEe}

\addplot[
	color=black,
	mark=*,
	]
	plot coordinates {
    (0, 0.60)
    (1, 0.56)
    (2, 0.54)
    (3, 0.53)
    (4, 0.49)
    (5, 0.47)
    (6, 0.44)
    (7, 0.40)
    (8, 0.36)
    (9, 0.30)
    
	};
\addlegendentry{DocNADE}

\addplot[
	color=green,
	mark=*,
	]
	plot coordinates {
    (0, 0.68)
    (1, 0.64)
    (2, 0.62)
    (3, 0.60)
    (4, 0.56)
    (5, 0.53)
    (6, 0.50)
    (7, 0.45)
    (8, 0.40)
    (9, 0.33)
	};
\addlegendentry{glove}

\end{axis}
\end{tikzpicture}%
\caption{{\bf IR:} TMNtitle} \label{IRTMNtitle}
\end{subfigure}\hspace*{\fill}%
~%
\begin{subfigure}{0.30\textwidth}
\centering
\begin{tikzpicture}[scale=0.65][baseline]
\begin{axis}[
    xlabel={\bf Fraction of Retrieved Documents (Recall)},
    ylabel={\bf Precision (\%)},
    xmin=0, xmax=9,
    ymin=0.39, ymax=0.81,
    /pgfplots/ytick={.40,.45,...,.81},
    xtick={0,1,2,3,4,5,6,7,8,9},
    xticklabels={0.0001, 0.0005, 0.001, 0.002, 0.005, 0.01, 0.02, 0.05, 0.1, 0.2},
    x tick label style={rotate=48,anchor=east},
    legend pos=south west,
    ymajorgrids=true, xmajorgrids=true,
    grid style=dashed,
]

\addplot[
	color=cyan,
	mark=triangle,
	]
	plot coordinates {
    (0, 0.76)
    (1, 0.73)
    (2, 0.72)
    (3, 0.71)
    (4, 0.69)
    (5, 0.68)
    (6, 0.66)
    (7, 0.62)
    (8, 0.57)
    (9, 0.49)
	};
\addlegendentry{iDeepDNEe}

\addplot[
	color=orange,
	mark=*,
	]
	plot coordinates {
    (0, 0.68)
    (1, 0.65)
    (2, 0.65)
    (3, 0.64)
    (4, 0.62)
    (5, 0.61)
    (6, 0.59)
    (7, 0.56)
    (8, 0.53)
    (9, 0.47)
	};
\addlegendentry{DeepDNE}

\addplot[
	color=blue,
	mark=square,
	]
	plot coordinates {
    (0, 0.77)
    (1, 0.75)
    (2, 0.73)
    (3, 0.72)
    (4, 0.71)
    (5, 0.69)
    (6, 0.67)
    (7, 0.63)
    (8, 0.59)
    (9, 0.51)
	};
\addlegendentry{iDocNADEe}

\addplot[
	color=red,
	mark=square,
	]
	plot coordinates {
   (0, 0.75)
    (1, 0.73)
    (2, 0.71)
    (3, 0.70)
    (4, 0.68)
    (5, 0.67)
    (6, 0.64)
    (7, 0.61)
    (8, 0.56)
    (9, 0.49)
	};
\addlegendentry{iDocNADE}

\addplot[
	color=violet,
	mark=square,
	]
	plot coordinates {
   (0, 0.78)
    (1, 0.74)
    (2, 0.73)
    (3, 0.71)
    (4, 0.69)
    (5, 0.68)
    (6, 0.66)
    (7, 0.62)
    (8, 0.58)
    (9, 0.50)
	};
\addlegendentry{DocNADEe}

\addplot[
	color=black,
	mark=*,
	]
	plot coordinates {
    (0, 0.73)
    (1, 0.70)
    (2, 0.69)
    (3, 0.67)
    (4, 0.65)
    (5, 0.64)
    (6, 0.62)
    (7, 0.58)
    (8, 0.54)
    (9, 0.46)
	};
\addlegendentry{DocNADE}

\addplot[
	color=green,
	mark=*,
	]
	plot coordinates {
    (0, 0.75)
    (1, 0.71)
    (2, 0.69)
    (3, 0.67)
    (4, 0.64)
    (5, 0.61)
    (6, 0.58)
    (7, 0.53)
    (8, 0.48)
    (9, 0.41)
	};
\addlegendentry{glove}

\end{axis}
\end{tikzpicture}%
\caption{{\bf IR:} AGnewstitle} \label{IRAGnewstitle}
\end{subfigure}\hspace*{\fill}%
~%
\begin{subfigure}{0.30\textwidth}
\centering
\begin{tikzpicture}[scale=0.65][baseline]
\begin{axis}[
    xlabel={\bf Fraction of Retrieved Documents (Recall)},
    ylabel={\bf Precision (\%)},
    xmin=0, xmax=8,
    ymin=0.11, ymax=0.50,
    /pgfplots/ytick={.12,.16,...,.50},
    xtick={0,1,2,3,4,5,6,7,8},
    xticklabels={0.0005, 0.001, 0.002, 0.005, 0.01, 0.02, 0.05, 0.1, 0.2},
    x tick label style={rotate=48,anchor=east},
    legend pos=north east,
    ymajorgrids=true, xmajorgrids=true,
    grid style=dashed,
]

\addplot[
	color=cyan,
	mark=triangle,
	]
	plot coordinates {
    (0, 0.39)
    (1, 0.38)
    (2, 0.34)
    (3, 0.32)
    (4, 0.29)
    (5, 0.26)
    (6, 0.22)
    (7, 0.18)
    (8, 0.15)
	};
\addlegendentry{iDeepDNEe}

\addplot[
	color=orange,
	mark=*,
	]
	plot coordinates {
    (0, 0.33)
    (1, 0.33)
    (2, 0.30)
    (3, 0.26)
    (4, 0.23)
    (5, 0.21)
    (6, 0.17)
    (7, 0.15)
    (8, 0.12)
	};
\addlegendentry{DeepDNE}

\addplot[
	color=blue,
	mark=square,
	]
	plot coordinates {
    (0, 0.46)
    (1, 0.46)
    (2, 0.42)
    (3, 0.37)
    (4, 0.33)
    (5, 0.28)
    (6, 0.23)
    (7, 0.19)
    (8, 0.15)
	};
\addlegendentry{iDocNADEe}

\addplot[
	color=red,
	mark=square,
	]
	plot coordinates {
    (0, 0.44)
    (1, 0.44)
    (2, 0.39)
    (3, 0.33)
    (4, 0.30)
    (5, 0.26)
    (6, 0.22)
    (7, 0.18)
    (8, 0.14)
	};
\addlegendentry{iDocNADE}

\addplot[
	color=violet,
	mark=square,
	]
	plot coordinates {
    (0, 0.47)
    (1, 0.47)
    (2, 0.42)
    (3, 0.37)
    (4, 0.32)
    (5, 0.28)
    (6, 0.23)
    (7, 0.19)
    (8, 0.14)
	};
\addlegendentry{DocNADEe}

\addplot[
	color=black,
	mark=*,
	]
	plot coordinates {
    (0, 0.42)
    (1, 0.42)
    (2, 0.40)
    (3, 0.34)
    (4, 0.31)
    (5, 0.25)
    (6, 0.21)
    (7, 0.18)
    (8, 0.14)
	};
\addlegendentry{DocNADE}

\addplot[
	color=green,
	mark=*,
	]
	plot coordinates {
    (0, 0.38)
    (1, 0.38)
    (2, 0.37)
    (3, 0.32)
    (4, 0.28)
    (5, 0.23)
    (6, 0.18)
    (7, 0.15)
    (8, 0.12)
	};
\addlegendentry{glove}


\end{axis}
\end{tikzpicture}%
\caption{{\bf IR:} 20NSshort} \label{IR20NSshort}
\end{subfigure}\hspace*{\fill}%
~%

\begin{subfigure}{0.30\textwidth}
\centering
\begin{tikzpicture}[scale=0.65][baseline]
\begin{axis}[
    xlabel={\bf Fraction of Retrieved Documents (Recall)},
    ylabel={\bf Precision (\%)},
    xmin=0, xmax=10,
    ymin=0.05, ymax=0.43,
    /pgfplots/ytick={.10,.14,...,.43},
    xtick={0,1,2,3,4,5,6,7,8,9,10}, 
    xticklabels={0.0001, 0.0005, 0.001, 0.002, 0.005, 0.01, 0.02, 0.05, 0.1, 0.2, 0.3}, 
    x tick label style={rotate=48,anchor=east},
    legend pos=south west,
    ymajorgrids=true, xmajorgrids=true,
    grid style=dashed,
]

\addplot[
	color=cyan,
	mark=triangle,
	]
	plot coordinates {
    (0, 0.41)
    (1, 0.39)
    (2, 0.38)
    (3, 0.36)
    (4, 0.33)
    (5, 0.31)
    (6, 0.29)
    (7, 0.23)
    (8, 0.18)
    (9, 0.14)
    (10, 0.11)
	};
\addlegendentry{iDeepDNEe}

\addplot[
	color=orange,
	mark=*,
	]
	plot coordinates {
    (0, 0.37)
    (1, 0.35)
    (2, 0.34)
    (3, 0.33)
    (4, 0.31)
    (5, 0.29)
    (6, 0.25)
    (7, 0.22)
    (8, 0.18)
    (9, 0.13)
    (10, 0.11)
	};
\addlegendentry{DeepDNE}

\addplot[
	color=blue,
	mark=square,
	]
	plot coordinates {
    (0, 0.40)
    (1, 0.38)
    (2, 0.35)
    (3, 0.33)
    (4, 0.30)
    (5, 0.28)
    (6, 0.25)
    (7, 0.19)
    (8, 0.15)
    (9, 0.12)
    (10, 0.10)
	};
\addlegendentry{iDocNADEe}

\addplot[
	color=red,
	mark=square,
	]
	plot coordinates {
    (0, 0.39)
    (1, 0.36)
    (2, 0.34)
    (3, 0.32)
    (4, 0.29)
    (5, 0.27)
    (6, 0.24)
    (7, 0.20)
    (8, 0.16)
    (9, 0.12)
    (10, 0.10)
	};
\addlegendentry{iDocNADE}

\addplot[
	color=violet,
	mark=square,
	]
	plot coordinates {
    (0, 0.41)
    (1, 0.38)
    (2, 0.36)
    (3, 0.34)
    (4, 0.31)
    (5, 0.28)
    (6, 0.25)
    (7, 0.19)
    (8, 0.15)
    (9, 0.12)
    (10, 0.10)
	};
\addlegendentry{DocNADEe}

\addplot[
	color=black,
	mark=*,
	]
	plot coordinates {
    (0, 0.27)
    (1, 0.26)
    (2, 0.25)
    (3, 0.25)
    (4, 0.23)
    (5, 0.22)
    (6, 0.20)
    (7, 0.17)
    (8, 0.14)
    (9, 0.11)
    (10, 0.09)
	};
\addlegendentry{DocNADE}

\addplot[
	color=green,
	mark=*,
	]
	plot coordinates {
    (0, 0.32)
    (1, 0.31)
    (2, 0.28)
    (3, 0.26)
    (4, 0.23)
    (5, 0.20)
    (6, 0.17)
    (7, 0.13)
    (8, 0.11)
    (9, 0.08)
    (10, 0.07)
	};
\addlegendentry{glove}

\end{axis}
\end{tikzpicture}%
\caption{{\bf IR:} 20NS} \label{IR20newsgroup}
\end{subfigure}\hspace*{\fill}%
~%
\begin{subfigure}{0.30\textwidth}
\centering
\begin{tikzpicture}[scale=0.65][baseline]
\begin{axis}[
    xlabel={\bf Fraction of Retrieved Documents (Recall)},
    ylabel={\bf Precision (\%)},
    xmin=0, xmax=9,
    ymin=0.63, ymax=0.95,
   /pgfplots/ytick={.64,.67,...,.94},
    xtick={0,1,2,3,4,5,6,7,8,9},
    xticklabels={0.0005, 0.001, 0.002, 0.005, 0.01, 0.02, 0.05, 0.1, 0.2, 0.3}, 
    x tick label style={rotate=45,anchor=east},
    legend pos=south west,
    ymajorgrids=true, xmajorgrids=true,
    grid style=dashed,
]

\addplot[
	color=cyan,
	mark=triangle,
	]
	plot coordinates {
    (0, 0.93)
    (1, 0.93)
    (2, 0.92)
    (3, 0.92)
    (4, 0.91)
    (5, 0.90)
    (6, 0.86)
    (7, 0.82)
    (8, 0.77)
    (9, 0.73)
    (10, 0.59)
    (11, 0.43)
	};
\addlegendentry{iDeepDNEe}

\addplot[
	color=orange,
	mark=*,
	]
	plot coordinates {
    (0, 0.91)
    (1, 0.91)
    (2, 0.90)
    (3, 0.90)
    (4, 0.89)
    (5, 0.85)
    (6, 0.84)
    (7, 0.80)
    (8, 0.75)
    (9, 0.70)
    (10, 0.50)
    (11, 0.42)
	};
\addlegendentry{DeepDNE}

\addplot[
	color=blue,
	mark=square,
	]
	plot coordinates {
    (0, 0.94)
    (1, 0.93)
    (2, 0.92)
    (3, 0.91)
    (4, 0.90)
    (5, 0.87)
    (6, 0.83)
    (7, 0.79)
    (8, 0.74)
    (9, 0.71)
	};
\addlegendentry{iDocNADEe}

\addplot[
	color=red,
	mark=square,
	]
	plot coordinates {
    (0, 0.94)
    (1, 0.93)
    (2, 0.93)
    (3, 0.92)
    (4, 0.90)
    (5, 0.88)
    (6, 0.83)
    (7, 0.79)
    (8, 0.75)
    (9, 0.71)
	};
\addlegendentry{iDocNADE}

\addplot[
	color=violet,
	mark=square,
	]
	plot coordinates {
    (0, 0.93)
    (1, 0.92)
    (2, 0.92)
    (3, 0.91)
    (4, 0.89)
    (5, 0.87)
    (6, 0.82)
    (7, 0.78)
    (8, 0.74)
    (9, 0.70)
	};
\addlegendentry{DocNADEe}

\addplot[
	color=black,
	mark=*,
	]
	plot coordinates {
    (0, 0.91)
    (1, 0.90)
    (2, 0.89)
    (3, 0.88)
    (4, 0.86)
    (5, 0.83)
    (6, 0.78)
    (7, 0.74)
    (8, 0.69)
    (9, 0.65)
	};
\addlegendentry{DocNADE}

\addplot[
	color=green,
	mark=*,
	]
	plot coordinates {
    (0, 0.92)
    (1, 0.92)
    (2, 0.91)
    (3, 0.89)
    (4, 0.87)
    (5, 0.84)
    (6, 0.79)
    (7, 0.75)
    (8, 0.69)
    (9, 0.65)
	};
\addlegendentry{glove}

\end{axis}
\end{tikzpicture}%
\caption{{\bf IR:} Reuters8} \label{IRReuters8}
\end{subfigure}\hspace*{\fill}%
~%
\begin{subfigure}{0.30\textwidth}
\centering
\begin{tikzpicture}[scale=0.65][baseline]
\begin{axis}[
    xlabel={\bf Fraction of Retrieved Documents  (Recall)},
    ylabel={\bf Precision (\%)},
    xmin=0, xmax=9,
    ymin=0.49, ymax=0.87,
    /pgfplots/ytick={.50,.55,...,.87},
    xtick={0,1,2,3,4,5,6,7,8,9}, 
    xticklabels={0.0001, 0.0005, 0.001, 0.002, 0.005, 0.01, 0.02, 0.05, 0.1, 0.2}, 
    x tick label style={rotate=48,anchor=east},
    legend pos=south west,
    ymajorgrids=true, xmajorgrids=true,
    grid style=dashed,
]

\addplot[
	color=cyan,
	mark=triangle,
	]
	plot coordinates {
    (0, 0.86)
    (1, 0.84)
    (2, 0.84)
    (3, 0.83)
    (4, 0.82)
    (5, 0.81)
    (6, 0.80)
    (7, 0.75)
    (8, 0.70)
    (9, 0.61)
	};
\addlegendentry{iDeepDNEe}

\addplot[
	color=orange,
	mark=*,
	]
	plot coordinates {
    (0, 0.82)
    (1, 0.81)
    (2, 0.80)
    (3, 0.80)
    (4, 0.79)
    (5, 0.78)
    (6, 0.75)
    (7, 0.73)
    (8, 0.69)
	};
\addlegendentry{DeepDNE}

\addplot[
	color=blue,
	mark=square,
	]
	plot coordinates {
    (0, 0.86)
    (1, 0.84)
    (2, 0.83)
    (3, 0.82)
    (4, 0.82)
    (5, 0.805)
    (6, 0.79)
    (7, 0.74)
    (8, 0.70)
    (9, 0.62)
	};
\addlegendentry{iDocNADEe}

\addplot[
	color=red,
	mark=square,
	]
	plot coordinates {
   (0, 0.86)
    (1, 0.85)
    (2, 0.84)
    (3, 0.83)
    (4, 0.81)
    (5, 0.79)
    (6, 0.78)
    (7, 0.73)
    (8, 0.68)
    (9, 0.60)
	};
\addlegendentry{iDocNADE}

\addplot[
	color=violet,
	mark=square,
	]
	plot coordinates {
   (0, 0.85)
    (1, 0.83)
    (2, 0.82)
    (3, 0.81)
    (4, 0.79)
    (5, 0.78)
    (6, 0.76)
    (7, 0.72)
    (8, 0.67)
    (9, 0.59)
	};
\addlegendentry{DocNADEe}

\addplot[
	color=black,
	mark=*,
	]
	plot coordinates {
    (0, 0.85)
    (1, 0.83)
    (2, 0.82)
    (3, 0.80)
    (4, 0.78)
    (5, 0.76)
    (6, 0.72)
    (7, 0.67)
    (8, 0.60)
    (9, 0.51)
	};
\addlegendentry{DocNADE}

\addplot[
	color=green,
	mark=*,
	]
	plot coordinates {
    (0, 0.84)
    (1, 0.81)
    (2, 0.79)
    (3, 0.78)
    (4, 0.76)
    (5, 0.74)
    (6, 0.72)
    (7, 0.68)
    (8, 0.63)
    (9, 0.54)
	};
\addlegendentry{glove}

\end{axis}
\end{tikzpicture}%
\caption{{\bf IR:} AGnews} \label{IRAGnews}
\end{subfigure}\hspace*{\fill}%
\caption{
Document retrieval performance (IR-precision) on 
3 short-text and 3 long-text datasets at different retrieval fractions
}
\label{fig:docretrieval}
\end{figure*}
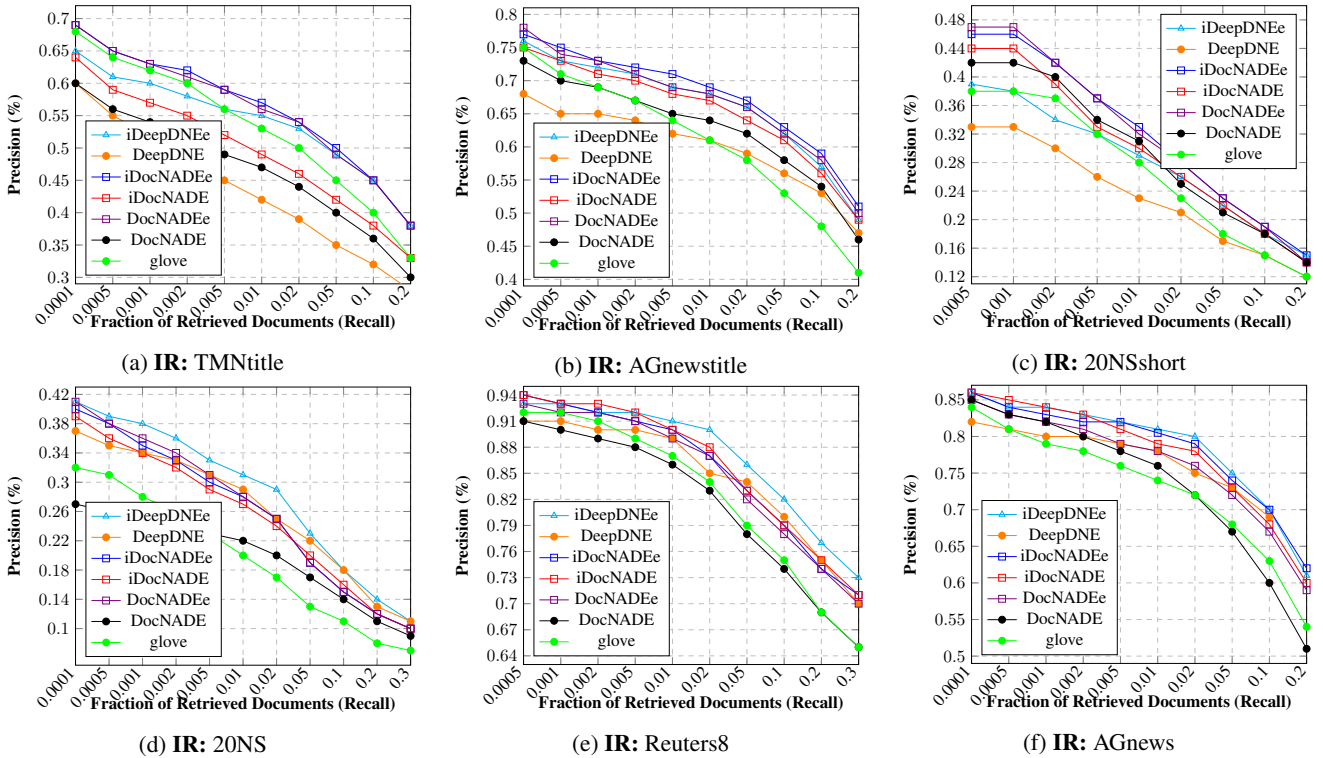
\makeatletter

{\bf Quantitative:} 
Table \ref{datastatisticsPPLIRshortlongtext}  shows the average held-out perplexity ($PPL$) per word as, 
$ PPL = \exp \big( - \frac{1}{N} \sum_{t=1}^{N} \frac{1}{|{\bf v}^t|} \log p({\bf v}^{t}) \big)$  
where $N$  and $|{\bf v}^t|$ are the total number of documents and words in a document ${\bf v}^{t}$.   
To compute PPL, the log-likelihood of the document ${\bf v}^{t}$, i.e., $\log p({\bf v}^{t})$, is obtained by 
$\mathcal{L}^{DocNADE}$ (eqn. \ref{eq:DocNADEloglikelihood}) in the DocNADE (forward only) variants, while 
we average PPL scores from the forward and backward networks of the iDocNADE variants. 

Table \ref{datastatisticsPPLIRshortlongtext} shows that the proposed models achieve lower perplexity for both the short-text (\underline{$413$} vs $435$) 
and long-text ({\bf $393$} vs $416$) datasets than {\it baseline} DocNADE with full-softmax (or tree-softmax). 
In total, we show a gain of $5.2$\% (404 vs 426) in PPL score on an average over the 15 datasets. 

Table \ref{PPLIRdeepshortlongtext} illustrates the generalization performance of deep variants, where the proposed extensions outperform the DeepDNE for both short-text and long-text datasets.   
We report a gain of 10.7\% ($402$ vs $450$)  in PPL due to iDeepDNEe over the baseline DeepDNE, on an average over 11 datasets.  

{\bf Inspection:} We quantify the use of context information in learning informed document representations. 
For 20NS dataset, 
we randomly select 50 held-out documents from its test set
 and compare (Figure \ref{PPL: 20NewsGroup}) 
the {\it PPL}  
for each of the held-out documents under the learned 200-dimensional DocNADE and iDocNADE.   
Observe that iDocNADE achieves lower {\it PPL} for the majority of the documents.  
The {\it filled} circle(s) 
points to the document for which {\it PPL} differs by a maximum between iDocNADE and DocNADE.  
We select the corresponding document and compute the negative log-likelihood ({\it NLL}) for every word.  
Figure \ref{NLL20newsgroup} 
shows that the {\it NLL}  for the majority of the words is lower (better) in iDocNADE than DocNADE. 
See the {\it supplementary material} 
for the raw text of the selected documents.

\subsubsection{Interpretability (Topic Coherence)}
Beyond PPL, 
we compute topic coherence 
\cite{Chang:82,Newman:82,Das:82,Gupta:85} 
to assess the meaningfulness of the 
underlying topics captured. 
We choose the coherence measure proposed by \citeauthor{Michael:82}  \shortcite{Michael:82}   
that identifies context features for each topic word using a sliding window over the reference corpus. The higher scores imply more coherent topics.

{\bf Quantitative:} We use gensim module 
({\it coherence type} = $c\_v$) to estimate coherence 
for each of the 200 topics (top 10 and 20 words). 
Table \ref{COHshortlongtext} shows average coherence over 200 topics using short-text and long-text datasets,  
where the high scores for long-text in iDocNADE ($.636$ vs $.602$) suggest that the contextual information helps in generating more coherent topics than DocNADE.    
On top, the introduction of embeddings, i.e., iDocNADEe for short-text boosts ($.847$ vs $.839$)  topic coherence.  
{\bf Qualitative:} Table \ref{topiccoherence} illustrates example topics each with a coherence score.

\enote{hs}{is $D$ used for two different purposes
(length document / scaling factor)
or is it
the same? (e.g., do you use something like average length of
a document for $D$)}
\enote{pg}{$D$ plays role for scaling and document length. Both are same.}

\begin{table}[t]
\center
\renewcommand*{\arraystretch}{1.2}
\resizebox{.43\textwidth}{!}{
\begin{tabular}{r|rr|rr|rr|rr}
\multirow{2}{*}{\bf data}     & \multicolumn{2}{c|}{{\bf DeepDNE}}  & \multicolumn{2}{c|}{\bf iDeepDNE}     & \multicolumn{2}{c|}{\bf  DeepDNEe}  & \multicolumn{2}{c}{\bf  iDeepDNEe}  \\ \cline{2-9}

                            & {\bf PPL}   & {\bf IR}        & {\bf PPL}   & {\bf IR}        & {\bf PPL}   & {\bf IR}         & {\bf PPL}   & {\bf IR}     \\ \hline

20NSshort              &    917    &    .21               &    841    &    .22               &    \underline{827}    &    .25               &    830    &    {\bf .26}          \\


TREC6                  &    114    &    .50                &    69    &    .52            &    69    &    {\bf .55}                 &   \underline{68}     &    {\bf .55}          \\

R21578title           &    253    &    .50               &    231    &    .52               &    236    &    {\bf .63}               &    \underline{230}    &    .61          \\

Subjectivity           &    428    &    .77               &    393    &    .77               &    \underline{392}    &    .81               &    \underline{392}    &    {\bf .82}          \\

Polarity                 &    408    &    .51               &    385    &    .51               &    \underline{383}    &    {\bf .55}               &    387    &    .53          \\

TMN                 &    681    &    .60               &    624    &    .62               &    627    &    .63               &    \underline{623}    &    {\bf .66}          \\


\hline


{\bf Avg (short)}  &    467    &    .51               &    424    &    .53               &    422    &   {\bf  .57}               &    \underline{421}    &   {\bf .57}          \\\hline \hline

Reuters8            &    216    &    .85               &    192    &    .89               &    \underline{191}    &    .88               &    \underline{191}    &    {\bf .90}          \\

20NS                 &    551    &    .25               &    \underline{504}    &    .28               &    \underline{504}    &    {\bf .29}               &    506    &    {\bf .29}          \\

R21578             &    318    &    .71               &    299    &    {\bf .73}               &    \underline{297}    &    .72               &    298    &    {\bf .73}          \\

AGNews            &    572    &    .75               &    441    &    .77                &   441     &    .75               &    \underline{440}    &    {\bf .80}          \\

RCV1V2            &    489    &    .86               &    464    &    .88               &    466    &    {\bf .89}               &    \underline{462}    &   {\bf .89}          \\ \hline

{\bf Avg (long)}  &   429    &    .68               &    380    &    .71               &   \underline{379}    &    .71               &    \underline{379}    &   {\bf  .72}          \\  \hline \hline 

{\bf Avg (all)}     &    450    &    .59               &    404    &    .61               &  403    &    .63               &    \underline{402}    &    {\bf .64}          \\                 

\end{tabular}}
\caption{Deep Variants (+ Full-softmax) with T200: PPL and IR (i.e, IR-precision) for {\bf short} and {\bf long} text datasets.}
\label{PPLIRdeepshortlongtext}
\end{table}
\subsubsection{Applicability (Document Retrieval)}
To evaluate the quality of the learned representations, 
we perform a document retrieval task using the 15 datasets and their label information.
We use the experimental setup similar to  
\citeauthor{HugoJMLR:82} \shortcite{HugoJMLR:82}, 
where all test documents are treated as queries to retrieve a fraction of the closest documents in
the original training set using cosine similarity measure between their representations (eqn. \ref{eq:documentrepiDocNADE} in iDocNADE 
and $\overrightarrow{\bf h}_D$ in DocNADE).  
To compute retrieval precision for each fraction (e.g., $0.0001$, $0.005$, $0.01$, $0.02$, $0.05$, $0.1$, $0.2$, etc.), 
we average the number of retrieved training documents with the same label as the query. 
For multi-label datasets, we average the precision scores over multiple labels for each query.  
Since 
\citeauthor{Salakhutdinov:82} \shortcite{Salakhutdinov:82} 
and  
\citeauthor{HugoJMLR:82} \shortcite{HugoJMLR:82}  
showed that RSM and DocNADE strictly outperform 
LDA on this task, we only compare DocNADE and its proposed extensions.    

Table \ref{datastatisticsPPLIRshortlongtext} shows the IR-precision scores at retrieval fraction $0.02$. 
Observe that the introduction of both pre-trained embedding priors and contextual information  leads to improved performance on the IR task for short-text and long-text datasets.    
We report a gain of $11.1$\% ($.60$ vs $.54$) in precision on an average over the 15 datasets, compared to DocNADE. 
On top, the deep variant i.e. iDeepDNEe (Table \ref{PPLIRdeepshortlongtext}) demonstrates a gain of 8.5\% ($.64$ vs $.59$) in precision over the 11 datasets, compared to DeepDNE.  
Figures (\ref{IRTMNtitle},  \ref{IRAGnewstitle},  \ref{IR20NSshort}) and (\ref{IR20newsgroup}, \ref{IRReuters8} and \ref{IRAGnews}) illustrate   
the average precision for the retrieval task on short-text  and long-text  datasets, respectively. 

\begin{table}[t]
\centering
\small
\renewcommand*{\arraystretch}{1.2}
\resizebox{.43\textwidth}{!}{
\setlength\tabcolsep{2.8pt}
\begin{tabular}{r|cc||cc|cc|cc}
\multirow{2}{*}{\bf model}  &   \multicolumn{2}{c||}{{\bf DocNADE}}  &   \multicolumn{2}{c|}{{\bf iDocNADE}}  &      \multicolumn{2}{c|}{{\bf DocNADEe}}  &   \multicolumn{2}{c}{{\bf iDocNADEe}} \\  

                         & W10      & W20           & W10     & W20    & W10          & W20      & W10   & W20 \\ \hline\hline

20NSshort           & .744     &     .849       & .748     & .852     &     .747       & .851     & .744    &     .849       \\

TREC6                & .746     &     .860       & .748     & .864     &     .753       & .858     & .752   &     .866         \\

R21578title        & .742     &     .845       & .748     & .855     &     .749       & .859     & .746   &     .856         \\


Polarity             & .730     &     .833       & .732     & .837     &     .734       & .839     & .738   &     .841    \\

TMNtitle            & .738     &     .840       & .744     & .848     &     .746       & .850     & .746   &     .850    \\ 

TMN                 & .709     &     .811       & .713     & .814     &     .717       & .818     &     .721       & .822   \\ \hline

{\bf Avg (short)}  & .734     &     .839       & .739     & .845     &     {\bf .742}    & .846   &     .741      &  {\bf .847}  \\  \hline \hline

20NSsmall            & .515     &     .629       & .564     & .669     &     .533       & .641     & .549   & .661                     \\

Reuters8            & .578     &     .665       & .564     & .657     &     .574       & .655     & .554   & .641                     \\

20NS                 &     .417       & .496     & .453     &     .531       & .385     & .458     & .417   & .490     \\

R21578              & .540     &     .570       & .548     & .640     &     .542       & .596     &     .551   & .663    \\

AGnews            & .718     &     .828       & .721     & .840     &     .677       & .739   &     .696       & .760    \\

RCV1V2         & .383     &     .426       & .428     & .480     &     .364       & .392     &     .420       & .463      \\
\hline

{\bf Avg (long)}   & .525     &     .602       & {\bf .546}     & {\bf .636}    & .513     & .580   &     .531      &     .613  

       
\end{tabular}}
\caption{Topic coherence with the top 10 (W10) and 20 (W20) words from topic models (T200).
Since, \cite{GuptatextTOvec:81} have shown that DocNADE outperforms both glove-DMM and glove-LDA, therefore DocNADE as the baseline.}
\label{COHshortlongtext}
\end{table}

\begin{table}[t]
\centering
\small
\renewcommand*{\arraystretch}{1.2}
\resizebox{.48\textwidth}{!}{
\setlength\tabcolsep{2.5pt}
\begin{tabular}{r|cc|cc||cc|cc|cc|cc}
\multirow{2}{*}{\bf data}   & \multicolumn{2}{c|}{{\bf glove}}    & \multicolumn{2}{c||}{{\bf doc2vec}}   & \multicolumn{2}{c|}{{\bf DocNADE}}  
 & \multicolumn{2}{c|}{\bf DocNADEe}       &  \multicolumn{2}{c|}{\bf iDocNADE}     
& \multicolumn{2}{c}{\bf iDocNADEe}  \\
\cline{2-13}
  
 & F1   & acc     & F1   & acc    & F1   & acc    & F1   & acc    & F1   & acc    & F1   & acc    \\ \hline 

20NSshort           &   .493      &     .520           &    .413   &     .457        &    .428       &  .474   &   .473       &  .529         &    .456      &   .491       &   .518       &     .535  \\

TREC6                &    .798     &       .810       &    .400   &     .512   &   .804        &  .822         &   .854       &    .856      &    .808      &  .812          &     .842     &   .844   \\

R21578title        &  .356       &    .695         &   .176    &   .505       &  .318         &   .653      &    .352      &     .693     &     .302     &  .665     &   .335             &     .700  \\

Subjectivity        &   .882      &     .882      &    .763      &  .763       &  .872         &     .872    &  .886        &   .886       &    .871      &     .871    &     .886           &   .886   \\

Polarity              &  .715       &    .715       &     .624     &  .624       &     .693      &    .693    &    .712      &   .712       &   .688       &   .688      &    .714            &     .714   \\

TMNtitle            &    .693     &   .727        & .582          &    .617   &    .624      & .667         &     .697     &    .732      &   .632       &        .675 &    .696            &     .731   \\

TMN                       &   .736      &    .755       &   .720       &    .751     &     .740      &    .778     &   .765       &   .801       &  .751         &     .790    &       .771         &    .805    \\

AGnewstitle      &  .814       &    .815        &   .513        &     .515     &    .812      &   .812     &    .829      &   .828       &    .819      &   .818      &     .829           &   .828   \\ \hline


{\bf Avg (short)} &  .685       &    .739       &      .523    &   .593      &   .661        &   .721      &    .696      &   {.755}       &    .666      &      .726   &       {\bf .700}         &   {\bf .756}       \\ \hline\hline  

Reuters8           &    .830       &     .950       &     .937      &   .852      &   .753        &     .931         &    .848      &    .956      &  .836        &    .957     &          .860      &    .960   \\

20NS                      &     .509    &   .525        &    .396      &   .409      &    .512       &      .535   &   .514       &  .540        &   .524        &  .548       &        .523        &  .544    \\

R21578                  &  .316       &    .703       &     .215     &   .622      &      .324     &   .716      &    .322      &   .721  &    .350            &   .710      &      .300          &      .722   \\


AGnews                &   .870      &    .871       &     .713     &   .711      &   .873        &   .876   &     .880     &   .880       &   .880       &   .880      &       .886         &   .886     \\

RCV1V2            &      .442   &      .368     &    .442      &   .341      &      .461     &     .438    &    .460        &     .457           &    .463          &   .452            &       .465              &   .454         \\
\hline

{\bf Avg (long)}           &  .593       &  .683         &      .540    &    .587     &     .584      &   .699      &  .605      &     .711     &   {\bf .611}       &    .710     &     { .607}          &  {\bf .713}        \\ \hline \hline

{\bf Avg (all)}             &  .650       &  .718         &    .530      &   .590      &    .631       &    .712     &   .661     &    .738      &    .645      &    .720     &       {\bf .664}         &    {\bf .740}      

\end{tabular}}
\caption{Text classification for short and long texts with T200 or word embedding dimension (Topic models with FS)}
\label{Classificationshortlongtext}
\end{table}

\begin{table}[t]
\centering
\renewcommand*{\arraystretch}{1.1}
\resizebox{0.42\textwidth}{!}{
\setlength\tabcolsep{2.5pt}
\begin{tabular}{c|c|c}
{\bf DocNADE}        & {\bf iDocNADE}       & {\bf DocNADEe}      \\ \hline

  beliefs, muslims,              &        scripture, atheists,                     &  atheists,     christianity,                 \\
  forward, alt,             &      sin, religions,    &           belief,    eternal,       \\
  islam, towards,              &    christianity, lord,                  &     atheism,    catholic,                       \\
  atheism, christianity,             &    bible, msg,           &        bible,    arguments,               \\
  hands, opinions             &   heaven, jesus           &         islam,    religions            \\ \hline 
      0.44         &        0.46                  &      \underline{0.52}                         
\end{tabular}}
\caption{Topics (top 10 words) of 20NS with coherence}
\label{topiccoherence}
\end{table}

\begin{table}[t]
\centering
\renewcommand*{\arraystretch}{1.15}
\resizebox{0.47\textwidth}{!}{
\setlength\tabcolsep{2.8pt}
\begin{tabular}{ccc|ccc|ccc|ccc}
\hline
\multicolumn{3}{c|}{\bf book}   & \multicolumn{3}{c|}{\bf jesus}   & \multicolumn{3}{c|}{\bf windows}   & \multicolumn{3}{c}{\bf gun}  \\ 
{\it neighbors} & {$s_i$}  & {$s_g$}  & {\it neighbors} & {$s_i$} & {$s_g$}  & {\it neighbors} & {$s_i$}    & {$s_g$}  & {\it neighbors} & {$s_i$}  & {$s_g$}     \\  \hline
books   &  .61    &    .84     &  christ      &   .86       &   .83         &  dos       &  .74         &    .34          &  guns   &   .72    &   .79          \\
reference   & .52    &  .51          &   god     &   .78       &  .63       &  files         &   .63         &  .36            &  firearms   &  .63       &  .63      \\
published   &    .46  &  .74       & christians      & .74         &    .49        &  version         &   .59        &    .43         &  criminal      &  .63   &  .33       \\
reading   &    .45   &    .54       & faith      &     .71       &   .51           & file         &  .59      &   .36       &  crime   &    .62     &   .42        \\
author  &   .44  &   .77         & bible      &   .71      &   .51      &    unix     &  .52          &   .47          &  police   & .61      &   .43     
\end{tabular}}
\caption{20NS dataset: The five nearest neighbors by iDocNADE.  $s_i$: Cosine similarity between the word vectors from iDocNADE, for instance vectors of {\it jesus} and {\it god}. 
$s_g$: Cosine similarity in embedding vectors from glove. 
}
\label{neighbors}
\end{table}

\subsubsection{Applicability (Text Categorization)}
Beyond the document retrieval, we perform text categorization to  measure the quality of word vectors learned in the topic models. 
We consider the same experimental setup as in the document retrieval task and extract the document representation (latent vector) of 200 dimension for each document (or text), 
learned during the training of DocNADE variants.  
To perform document categorization, we employ 
a logistic regression classifier 
with $L2$ regularization.  We also compute document
representations from pre-trained glove
\cite{pennington14glove2}
embedding matrix by summing the word vectors  
and compute classification performance. On top, we also extract document representation from doc2vec \cite{le14sentences}.

Table \ref{Classificationshortlongtext} shows that {\it glove} leads DocNADE in classification performance, suggesting a need for distributional priors. For short-text dataset, iDocNADEe (and  DocNADEe) 
outperforms  {\it glove} ($.700$ vs $.685$) and DocNADE ($.700$ vs $.661$) in F1. 
Overall, we report a gain of 5.2\% ($.664$ vs $.631$) in F1 due to iDocNADEe over DocNADE for classification on an average over 13 datasets.  


{\bf Inspection of Learned Representations}: 
To analyze the meaningful semantics captured, we perform a qualitative inspection of the learned representations by the topic models.   
Table \ref{topiccoherence} shows topics for 20NS dataset that could be interpreted as {\it religion}, 
which are (sub)categories in the data, confirming that meaningful topics are captured. 
Observe that DocNADEe extracts a more coherent topic. 

For word level inspection, 
we extract {\it word representations} using the columns $W_{:,v_i}$ as the vector (200 dimension) representation 
of each word $v_i$, learned by iDocNADE using 20NS dataset.  
Table \ref{neighbors} shows the five nearest neighbors of some selected words in this space 
and their corresponding  similarity scores. 
We also compare similarity in word vectors from iDocNADE and glove  
embeddings,  
 confirming that meaningful word representations are learned. 

\section{Conclusion}
We show 
that leveraging contextual information and introducing distributional priors via pre-trained word embeddings in our 
proposed topic models 
result in learning 
better word/document representation for short and long documents, and improve generalization, 
interpretability of topics and their applicability in text 
retrieval and classification. 

\section*{Acknowledgments}
We thank our colleagues  Bernt Andrassy,  Mark Buckley,  Stefan Langer, Ulli Waltinger and Subburam rajaram, and 
anonymous reviewers for their review comments. 
This research was supported by Bundeswirtschaftsministerium ({\tt bmwi.de}), grant 01MD15010A (Smart Data Web) 
at Siemens AG- CT Machine Intelligence, Munich Germany.

\bibliography{aaai19}
\bibliographystyle{aaai19}



\section*{Supplementary material}

# Supplementary Material

**AAAI Press**
Association for the Advancement of Artificial Intelligence
2275 East Bayshore Road, Suite 160
Palo Alto, California 94303

## Datasets

We use seven different datasets: (1) `20NSshort`: We take documents from 20NewsGroups data, with document size less (in terms of number of words) than 20. (2) `TREC6`: a set of questions (3) `Reuters21578title`: a collection of new stories from `nltk.corpus`. We take titles of the documents. (4) `Subjectivity`: sentiment analysis data. (5) `Polarity`: a collection of positive and negative snippets acquired from Rotten Tomatoes (6) `TMNtitle`: Titles of the Tag My News (TMN) news dataset. (7) `AGnewstitle`: Titles of the AGnews dataset. (8) `Reuters8`: a collection of news stories, processed and released by (9) `Reuters21578`: a collection of new stories from `nltk.corpus`. (10) `20NewsGroups`: a collection of news stories from `nltk.corpus`. (11) RCV1V2 (Reuters): www.ai.mit.edu/projects/jmlr/papers/volume5/lewis04a/lyrl2004_rcv1v2_README.htm (12) `Sixxx Requirement OBjects` (SiROBs): a collection of paragraphs extracted from industrial tender documents (our industrial corpus).

The SiROBs is our industrial corpus, extracted from industrial tender documents. The documents contain requirement specifications for an industrial project for example, *railway metro construction*. There are 22 types of requirements i.e. class labels (multi-class), where a requirement is a paragraph or collection of paragraphs within a document. We name the requirement as Requirement Objects (ROBs). Some of the requirement types are *project management*, *testing*, *legal*, *risk analysis*, *financial cost*, *technical requirement*, etc. We need to classify the requirements in the tender documents and assign each ROB to a relevant department(s). Therefore, we analyze such documents to automate decision making, tender comparison, similar tender as well as ROB retrieval and assigning ROBs to a relevant department(s) to optimize/expedite tender analysis. See some examples of ROBs from SiROBs corpus in Table 1.

## Experimental Setup and Hyperparameters for IR task

We set the maximum number of training passes to 1000, topics to 200 and the learning rate to 0.001 with $tanh$ hidden activation. We performed stochastic gradient descent based on the contrastive divergence approximation during 1000 training passes, with different learning rates. For model selection, we used the validation set as the query set and used the average precision at 0.02 retrieved documents as the performance measure. Note that the labels are not used during training. The class labels are only used to check if the retrieved documents have the same class label as the query document.

To perform document retrieval, we use the same train/development/test split of documents discussed in data statistics (experimental section) for all the datasets during learning. For model selection, we use the development set as the query set and use the average precision at 0.02 retrieved documents as the performance measure. We train DocNADE and iDocNADE models with 200 topics and perform stochastic gradient descent for 2000 training passes with different learning rates. Note that the labels are not used during training. The class labels are only used to check if the retrieved documents have the same class label as the query document. See Table 3 for the hyperparameters in the document retrieval task.

## Experimental Setup and Hyperparameters for doc2vec model

### Doc2Vec

We used gensim (https://github.com/RaRe-Technologies/gensim) to train Doc2Vec models for 12 datasets. Models were trained with distributed bag of words, for 1000 iterations using a window size of 5 and a vector size of 500.

### Classification task

We used the same split in training/development/test as for training the Doc2Vec models (also same split as in IR task) and trained a regularized logistic regression classifier on the inferred document vectors to predict class labels. In the case of multilabel datasets (`R21578`,`R21578title`, `RCV1V2`), we used a one-vs-all approach. Models were trained with a liblinear solver using L2 regularization and accuracy and macro-averaged F1 score were computed on the test set to quantify predictive power.

| |
|---|
| **Label:** *training* |
| Instructors shall have tertiary education and experience in the operation and maintenance of the equipment or sub-system of Plant. They shall be proficient in the use of the English language both written and oral. They shall be able to deliver instructions clearly and systematically. The curriculum vitae of the instructors shall be submitted for acceptance by the Engineer at least 8 weeks before the commencement of any training. |
| **Label:** *maintenance* |
| The Contractor shall provide experienced staff for 24 hours per Day, 7 Days per week, throughout the Year, for call out to carry out On-call Maintenance for the Signalling System. |
| **Label:** *cables* |
| Unless otherwise specified, this standard is applicable to all cables which include single and multi-core cables and wires, Local Area Network (LAN) cables and Fibre Optic (FO) cables. |
| **Label:** *installation* |
| The Contractor shall provide and permanently install the asset labels onto all equipment supplied under this Contract. The Contractor shall liaise and co-ordinate with the Engineer for the format and the content of the labels. The Contractor shall submit the final format and size of the labels as well as the installation layout of the labels on the respective equipment, to the Engineer for acceptance. |
| **Label:** *operations, interlocking* |
| It shall be possible to switch any station Interlocking capable of reversing the service into "Auto-Turnaround Operation". This facility once selected shall automatically route Trains into and out of these stations, independently of the ATS system. At stations where multiple platforms can be used to reverse the service it shall be possible to select one or both platforms for the service reversal. |

Table 1: SiROBs data: Example Documents (Requirement Objects) with their types (label).

| Hyperparameter | Search Space |
|---|---|
| learning rate | [0.001, 0.005, 0.01] |
| hidden units | [50, 200] |
| iterations | [2000] |
| activation function | sigmoid |
| scaling factor | [True, False] |

Table 2: Hyperparameters in Generalization evaluation in the DocNADE and iDocNADE for 50 and 200 topics. The underline signifies the optimal setting.

## Hyperparameters for PPL and IR

See Tables 2 and 3 for hyperparameters in generalization and retrieval tasks.

| Hyperparameter | Search Space |
|---|---|
| retrieval fraction | [0.02] |
| learning rate | [0.001, 0.01] |
| hidden units | [200] |
| activation function | [sigmoid, tanh] |
| iterations | [2000, 3000] |
| scaling factor | [True, False] |

Table 3: Hyperparameters in the Document Retrieval task. The underline signifies the optimal setting.

| (from 20NewsGroups data set) |
|---|
| The CD-ROM and manuals for the March beta – there is no X windows server there. |
| Will there be? Of course. (Even) if Microsoft supplies one with NT , other vendors will no doubt port their's to NT. |

Table 4: Raw text for the selected documents in generalization inspection from 20NewsGroups (20NS) dataset.

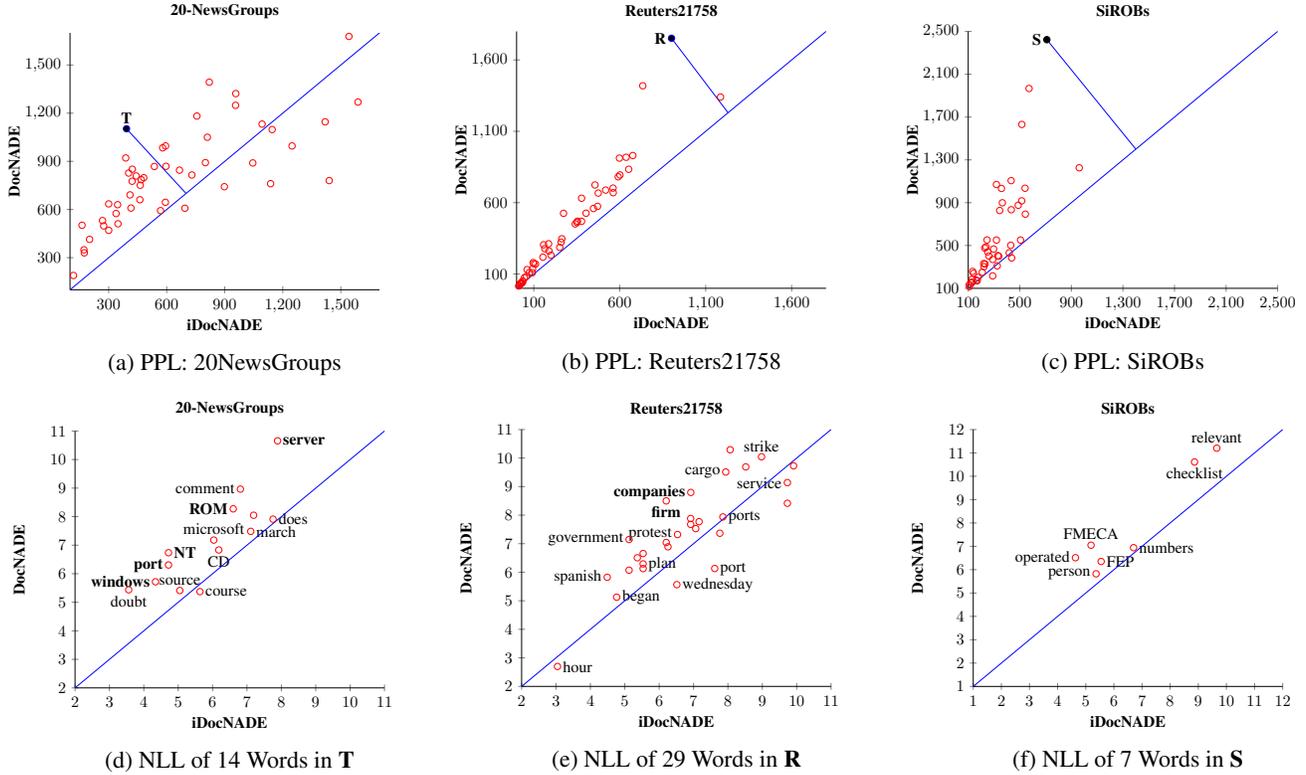

Figure 1: (a, b, c): PPL (200 topics) by iDocNADE and DocNADE for each of the 50 held-out documents. The *filled circle* and symbols (**T**, **R** and **S** point to the document for which *PPL* differs by maximum, each for 20NewsGroups, Reuters21758 and SiROBs datsets, respectively. (d, e, f): NLL of each of the words in documents marked by **T**, **R** and **S**, respectively due to iDocNADE and DocNADE.

| | PPL | | | |
|---|---|---|---|---|
| **Data** | **DocNADE** | | **iDocNADE** | |
| | $T50$ | $T200$ | $T50$ | $T200$ |
| *TREC6* | 42 | 42 | 39 | <u>39</u> |
| *Reuters8* | 178 | 172 | 162 | <u>152</u> |
| *Reuters21758* | 226 | 215 | 198 | <u>179</u> |
| *Polarity* | 310 | 311 | 294 | <u>292</u> |
| *20NewsGroups* | 864 | 830 | 836 | <u>812</u> |

Table 5: *PPL* and over 50 ($T50$) and 200 ($T200$) topics for DocNADE and iDocNADE using Tree-softmax. The <u>underline</u> indicates the best scores in PPL by iDocNADE.

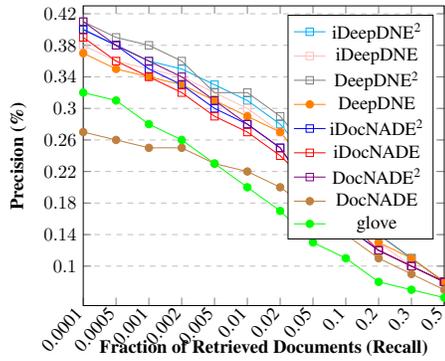
(a) **IR:** 20NS

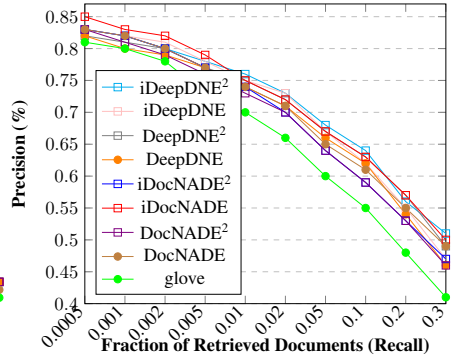
(b) **IR:** R21578

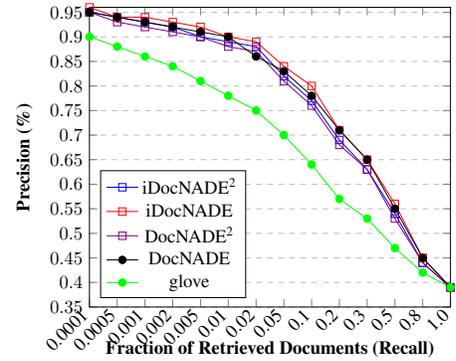
(c) **IR:** RCV1V2

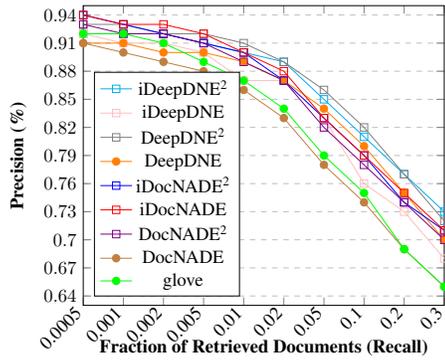
(d) **IR:** Reuters8

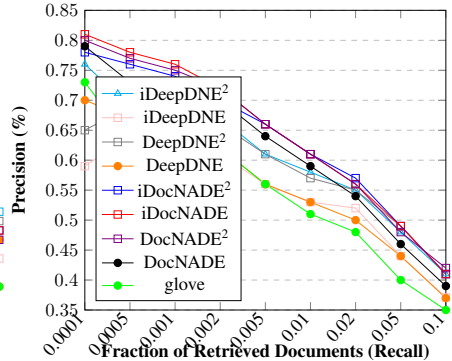
(e) **IR:** TREC6

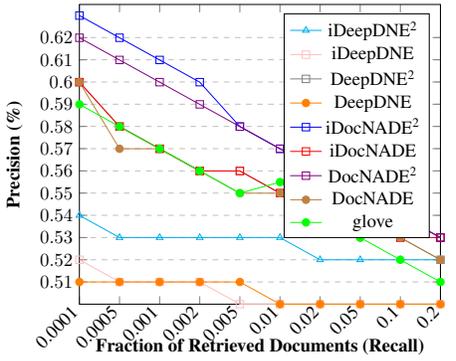
(f) **IR:** Polarity

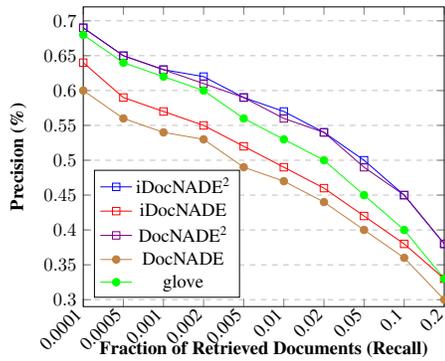
(g) **IR:** TMNtitle

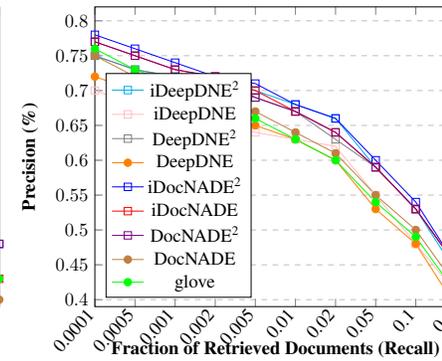
(h) **IR:** TMN

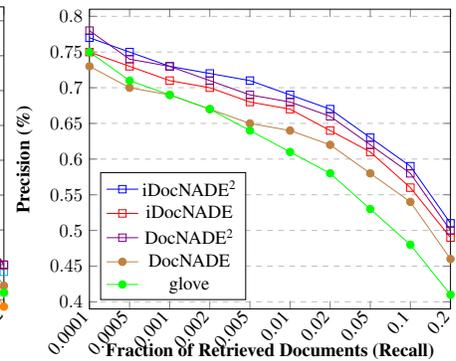
(i) **IR:** AGnewstitle

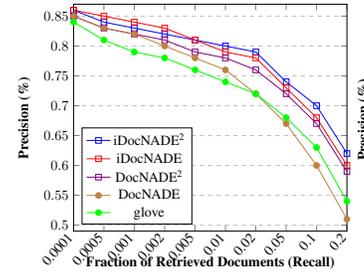
(j) **IR:** AGnews

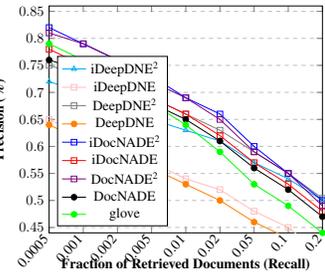
(k) **IR:** R21578title

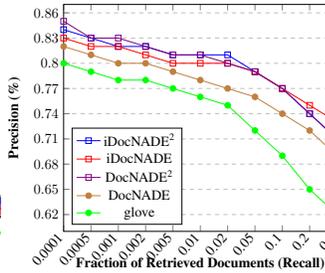
(l) **IR:** Subjectivity

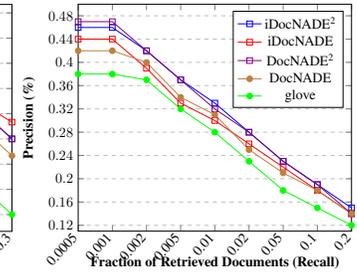
(m) **IR:** 20NSshort

Figure 2: Document retrieval performance (precision) at different retrieval fractions. Observe different y-axis scales.

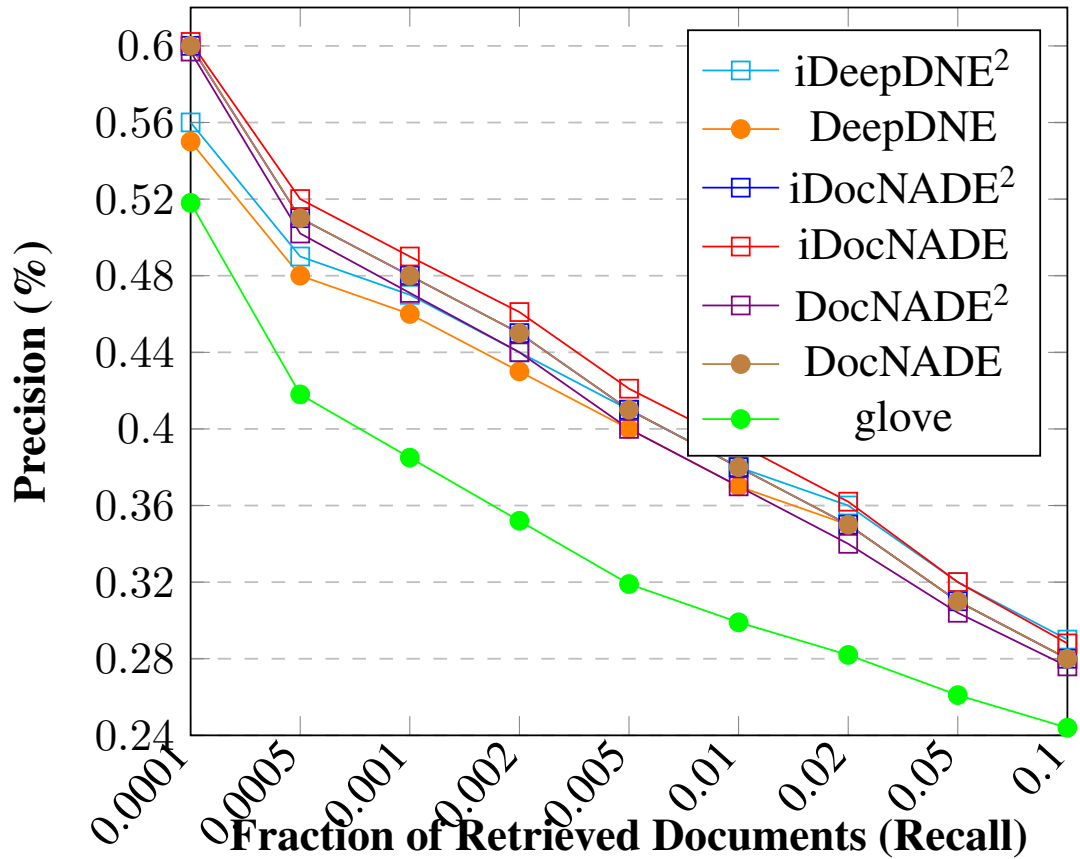

(a) **IR:** SiROBs

Figure 3: Document retrieval performance (precision) at different retrieval fractions. Observe different y-axis scales.



\end{document}